\documentclass{article}

\usepackage[english]{babel}

\usepackage[letterpaper,top=2cm,bottom=2cm,left=3cm,right=3cm,marginparwidth=1.75cm]{geometry}

\usepackage{amsmath}
\usepackage{graphicx}
\usepackage[colorlinks=true, allcolors=blue]{hyperref}

\usepackage{hyperref}       
\usepackage{url}            
\usepackage{booktabs}       
\usepackage{amsfonts}       
\usepackage{nicefrac}       
\usepackage{microtype}      
\usepackage{xcolor}         
\usepackage{authblk}

\usepackage{floatrow}
\newfloatcommand{capbtabbox}{table}[][\FBwidth]

\usepackage{booktabs} 
\usepackage{multicol}
\usepackage{blindtext}
\usepackage{xcolor}
\usepackage{amsthm}
\usepackage{graphicx}
\usepackage{subfigure}
\usepackage{multirow}
\usepackage{amssymb}
\usepackage{amsmath}
\usepackage{booktabs}
\usepackage{algorithm}
\usepackage{algorithmic}
\usepackage{color}
\usepackage{wrapfig}

\usepackage{bm}

\newcommand{\mba}{\bm{a}}
\newcommand{\mbb}{\bm{b}}

\newcommand{\mbx}{\bm{x}}
\newcommand{\mby}{\bm{y}}
\newcommand{\mbz}{\bm{z}}

\newcommand{\mbA}{\bm{A}}
\newcommand{\mbB}{\bm{B}}

\newcommand{\mbI}{\bm{I}}

\newcommand{\mbS}{\bm{S}}

\newcommand{\mbW}{\bm{W}}
\newcommand{\mbX}{\bm{X}}

\newcommand{\mbbh}{\hat{\mbb}}

\newcommand{\mc}[1]{\mathcal{#1}}
\newcommand{\mcH}{\mathcal{H}}
\newcommand{\mcN}{\mathcal{N}}

\newcommand{\mbbR}{\mathbb{R}}

\newcommand{\mbhx}{\bm{\hat{x}}}

\newcommand{\mbalpha}{\bm{\alpha}}

\newcommand{\mbepsilon}{\bm{\epsilon}}

\newcommand{\mbeta}{\bm{\eta}}

\newcommand{\mbOmega}{\bm{\Omega}}
\newcommand{\mbPhi}{\bm{\Phi}}

\newcommand{\mbPsi}{\bm{\Psi}}

\newcommand{\sign}{\textrm{sign}}
\newcommand{\Func}{\textrm{f}}
\newcommand{\FG}{\mathrm{G}}
\newcommand{\FD}{\mathrm{D}}

\newcommand{\Tr}{\textrm{Tr}}

\newcommand{\Dec}{\textrm{Dec}}
\newcommand{\Enc}{\textrm{Enc}}

\newcommand{\mcL}{\mathcal{L}}

\title{Sparse-Inductive Generative Adversarial Hashing for Nearest Neighbor Search}
\author[]{Hong LIU\thanks{lynnliu.xmu@gmail.com}}
\affil[]{National Institute of Informatics, Japan}
\date{}
\begin{document}
\maketitle

\begin{abstract}
Unsupervised hashing has received extensive research focus on the past decade, which typically aims at preserving a predefined metric (\emph{i.e.} Euclidean metric) in the Hamming space.
To this end, the encoding functions of the existing hashing are  typically quasi-isometric, which devote to reducing the quantization loss from the target metric space to the discrete Hamming space.
However, it is indeed problematic to directly minimize such error, since such mentioned two metric spaces are heterogeneous, and the quasi-isometric mapping is non-linear.
The former leads to inconsistent feature distributions, while the latter leads to problematic optimization issues.
In this paper, we propose a novel unsupervised hashing method, termed \textit{Sparsity-Induced Generative Adversarial Hashing} (SiGAH), to encode large-scale high-dimensional features into binary codes, which well solves the two problems through a generative adversarial training framework.
Instead of minimizing the quantization loss, our key innovation lies in enforcing the learned Hamming space to have similar data distribution to the target metric space via a generative model.
In particular, we formulate a ReLU-based neural network as a generator  to output binary codes and an MSE-loss based auto-encoder network as a discriminator, upon which a generative adversarial learning  is carried out to train hash functions.
Furthermore, to generate the synthetic features from the hash codes, a compressed sensing procedure is introduced into the generative model, which enforces the reconstruction boundary of binary codes to be consistent with that of original features.
Finally, such generative adversarial framework can be trained via the Adam optimizer. 
Experimental results on four benchmarks, \emph{i.e.}, Tiny100K, GIST1M, Deep1M, and MNIST, have shown that the proposed SiGAH has superior performance over the state-of-the-art approaches.
\end{abstract}

\section{Introduction}
%
%
%
%

Approximate nearest neighbor (ANN) search over large-scale image datasets has been a recent research hotspot in the fields of computer vision \cite{Wang:2015ev,Wang:2017gy}.
Given a query, ANN finds the most similar data using a predefined distance metric.
To this end, most recent works advocate the learning of binary codes, \emph{a.k.a.}, hashing, which achieves the best trade-off among  time, storage, and accuracy. 
In the past decade, various hashing schemes have been proposed, including, Iterative Quantization \cite{Gong2013IterativeQA}, Spherical hashing (SpH) \cite{Heo2015SphericalHB}, Anchor Graph Hashing \cite{liu2011hashing}, Kernel Supervised Hashing \cite{Liu2012SupervisedHW} and Ordinal Constrained Hashing (OCH) \cite{liu2018ordinal}.
\begin{figure*}[!t]\label{fig1}
\centering{
\includegraphics[width=0.8\linewidth]{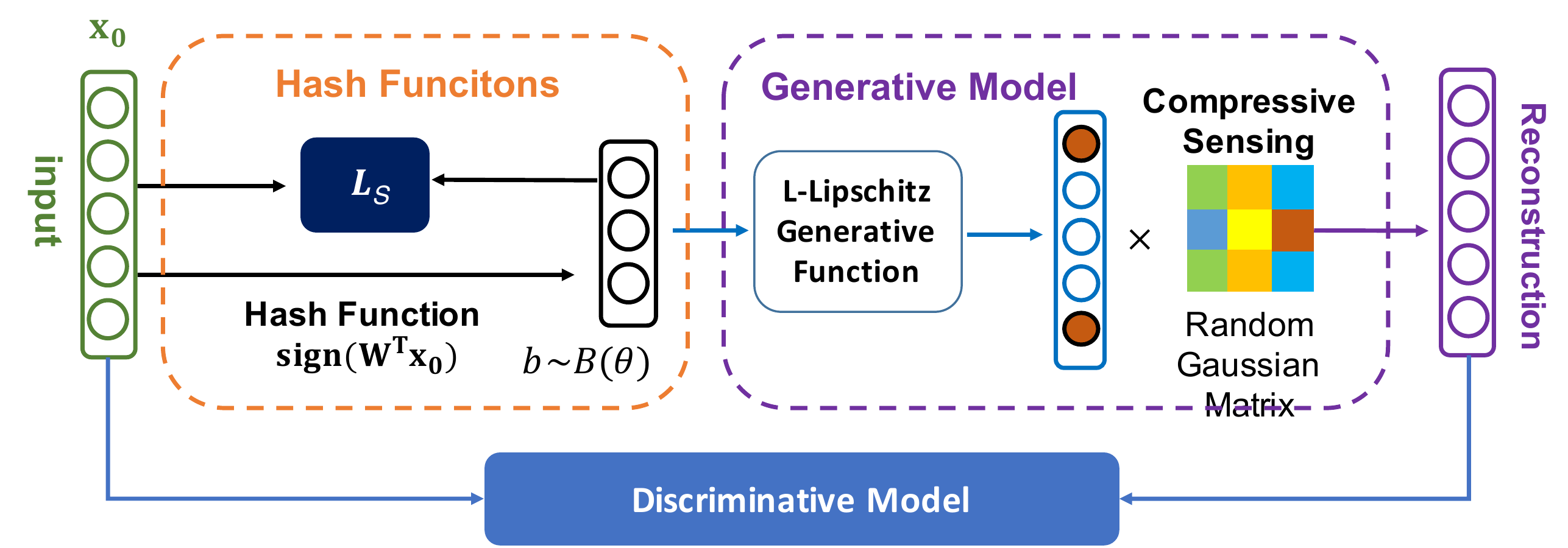}}
\caption{The framework of our proposed Sparsity-induced Generative Adversarial Hashing (SiGAH). 
SiGAH contains a generator and a discriminator, which forms a generative adversarial learning framework to help learn hash codes for image retrieval. 
In detail, we first design a sparsity-induced generator that aims to produce synthetic features by recovering the structured signal from the hash codes. 
Second, an energy-based discriminator is introduced to discriminate the distribution produced by the generator compared to the actual data distribution. 
Third, we combine the hash function learning with the above generative model, and  finally we use the GAN loss to train the discriminator and the generator to reach a Nash equilibrium. }
\end{figure*}

The earliest works in binary code learning aimed to find random projections or random permutations to produce hash functions \cite{Datar2004LocalitysensitiveHS,Broder2000MinWiseIP}. 
Later efforts have switched from such data-independent hashing to data-dependent methods, which can be subdivided into either supervised hashing \cite{Kulis2009LearningTH,Norouzi2011MinimalLH,Liu2012SupervisedHW,Shen2015SupervisedDH,Gui2017FastSD} or  unsupervised  hashing \cite{Gong2013IterativeQA,Heo2015SphericalHB,Liu2017OrdinalCB,Weiss2008SpectralH,Liu2011HashingWG,Liu2014DiscreteGH,Jiang2015ScalableGH,liu2018ordinal}. 

In this paper, we focus on unsupervised hashing, because semantic labeling typically requires a labor-intensive process that makes large-scale applications impossible.
Under this framework, the key challenge lies in deciding which information to preserve  from the original metric space when transferring to the Hamming space.  
Most existing hashing algorithms are devoted to preserving the original  pair-wise similarity in the Hamming space, \emph{i.e.}, forcing the Hamming distance between item pairs after binary coding to be consistent with that of the input space \cite{Wang:2017gy}.
This similarity is typically preserved by minimizing the quantization loss from the  original metric space to the target Hamming space, \emph{e.g.} Euclidean space \cite{Gong2013IterativeQA,Heo2015SphericalHB}.
However, there is the significant problem that the Euclidean space and Hamming space are heterogeneous.
As such, it is very challenging to determine the proper scales and transforms for comparing the distance/similarity between two spaces, especially in the case of unknown data distributions.   

To handle the distribution difference between the Hamming space and the original feature space, a new paradigm \cite{balu2014beyond} is to view data hashing as cross-space data reconstruction, which has shown that lower reconstruction error leads to better performance of the corresponding binary codes.
Therefore, several reconstruction based hashing methods \cite{CarreiraPerpin2015HashingWB,Dai2017StochasticGH} have shown significant merits in both theoretical advances and performance gains.
However, most of these follow a two-stage setting to encode the hash codes, \emph{i.e.}, dimensionality reduction and sign calculation. 
Such a two-stage strategy is quasi-isometric \cite{jacques2015quantized} and is hard to optimize \cite{balu2014beyond}. 
The targets were originally designed to reduce the quantization error \cite{Wang:2017gy}, rather than balance the distribution difference.

Following the recent advances in reconstruction-based hashing, we propose a novel hashing scheme, termed \textit{Sparsity-induced Generative Adversarial Hashing} (\textbf{SiGAH}), to handle the issues raised above.
In principle,  we aim to make the distribution of a recovered feature similar to the real feature.
This is done via a novel generative adversarial learning, \emph{i.e.}, by introducing a generative model to better approximate the input features, as well as introducing a discriminator to evaluate the approximation.
The framework of the proposed SiGAH is shown in Fig.\ref{fig1}. The three key innovations are presented as below:

\textit{In the generator part}, 
different from existing methods,  SiGAH attempts to minimize the  reconstruction error between an input feature and a synthetic feature, which serves as a regularization in the  formulation of  hash code learning.
To this end, we design a \emph{ sparsity-induced generative model} to produce synthetic features by recovering the structured signal from the hash codes, which differs from the  linear regression reconstruction used in the most recent work  \cite{CarreiraPerpin2015HashingWB}.
In particular, following the compressive sensing theory (\emph{i.e.},  recovering a structure vector $x\in \mathbb{R}^d$ after observing $r$-bit hash codes), the generation procedure is carried out in two steps: 
First, we learn a mapping from an $r$-bit hash codes $b \in \{0,1\}^r$ to a high-dimensional feature $G(b)\in \mathbb{R}^m~ (r < d < m )$, where $G(b)$ relies on a neural network with a sparsity constraint \cite{ogcMdT}.
Second, the original feature $x$ can be sufficiently recovered from the random Gaussian measurements $\mathbf{\Psi}$ and $G(b)$, where the recovered feature $\hat{x} = \mathbf{\Psi}G(b)$ is used as the synthetic feature.
Under the above setting, the generative model satisfies the $L$-Lipschitz function  \cite{jacques2015quantized}, which can force the reconstruction boundary of the recovered features to be consistent with the original ones.

\emph{In the discriminator part}, we design an energy-based model to match the  distribution produced by the generator to the actual data distribution. 
Inspired by \cite{zhao2016energy}, we use an auto-encoder architecture as the discriminator, in which the reconstruction error serves as the energy score.
Then, a margin loss is introduced to the discriminative model, which attributes higher energy to the synthetic features and lower energy to the real data.

\emph{In the hash learning part}, we feed the output of the hash layer (or hash function) to the input of the generator.
Therefore, the combined hash layer and the generator can be seen as a new generative model, \emph{i.e.}, the input layer is the original feature, the latent layer is the hash codes, and the output layer is the synthetic feature.
Moreover, we introduce a graph regularization to the latent layer of the new generator, so that the hash codes have similar representations in the Hamming space.
Similar to the vanilla generative adversarial network (GAN) \cite{goodfellow2014generative}, we use two different losses to iteratively train the discriminator and the generator to reach a Nash equilibrium.
Finally, we directly use a Stochastic Gradient Descent with the Adam optimizer \cite{Kingma2014AdamAM} to optimize the whole framework.

The proposed SiGAH method is compared against various state-of-the-art unsupervised hashing methods, including \cite{Datar2004LocalitysensitiveHS,Gong2013IterativeQA,Heo2015SphericalHB,Liu2017OrdinalCB,CarreiraPerpin2015HashingWB,Dai2017StochasticGH} on several widely-used ANN search benchmarks, \emph{i.e.}, Tiny100K, GIST1M, Deep1M, and MNIST.
Quantitative results demonstrate that our SiGAH outperforms the state-of-the-arts by  a significant margin.

The rest of this paper is organized as follows: 
related works and preliminaries are briefly introduced in section \ref{sec2}.
In section \ref{sec3}, we elaborate the proposed scheme of SiGAH and conduct an  algorithm analysis.
Section \ref{sec4} and section \ref{sec5} provide details of the network architecture,  the experimental results, and the corresponding analysis.
Finally, we conclude the overall paper in section \ref{sec6}.

\section{Related Work} \label{sec2}
In this section, we review several reconstruction-based hashing approaches. 
Essentially, quantization can be viewed as a reconstruction approach recovering individual  data items.
From this perspective, ITQ \cite{Gong2013IterativeQA} can be seen as a specific reconstruction-based hashing. 
It aims at minimizing the binary quantization loss, which  is essentially an $l_2$ reconstruction loss when the magnitude of the feature vectors is compatible with the radius of the binary cube.
Recently, \cite{balu2014beyond} have shown that the traditional ``projection and sign'' based hashing (termed quantization hashing) is sub-optimal when the projection is not orthogonal, based on which  a better encoding strategy is proposed  from the perspective of data reconstruction.
However, the proposed method requires that the hash bit to be larger than the dimension of the input feature, which is a strong restriction that  does not hold in many  applications.

For better feature reconstruction, the Auto-encoder (AE) model is a natural choice \cite{goodfellow2016deep}, which involves encoding-decoding of the input-output signals.
AE has also been extended to incorporate binary code learning, termed binary-AE \cite{CarreiraPerpin2015HashingWB}.
Given a dataset $\mathbf{X} = \{x_1, x_2,..., x_n\}$, binary-AE consists of two parts: an encoder model (hash function) $h(x_i)$  that maps a real-valued vector $x_i$ to a hidden representation $z_i$  with $r$ bits, $z_i\in \{0,1\}^{r}$, and a decoder model that produces a reconstruction $g(z)$.
As a result, binary-AE aims to minimize the following objective function, 
\begin{equation} \label{eq1}
    \min\limits_{h,g} \sum_{n=1}^{n} \|x_n - g(z_i)\|^2,~ z_i = h(x_i) \in \{0,1\}^r.
\end{equation}
The key issue of Eq.(\ref{eq1}) is that its reconstruction $g(z)$ is based on a simple linear regression, which cannot effectively preserve structural information in the original feature space.
Furthermore, optimizing Eq.(\ref{eq1}) is extremely difficult due to the binary constraints. 
Although auxiliary coordinates can help to optimize the model, the training efficiency is extremely high.

Recently, Stochastic Generative Hashing (SGH) \cite{Dai2017StochasticGH} was proposed which uses a generative model to reconstruct the original features.
SGH follows a Minimum Description Length principle to learn the hash functions.
This can also be viewed as a sort of auto-encoder hashing.
To avoid the difficulty in optimization caused  by binary constraints, SGH adopts a linear hash function   as the encoder, and proposes a Frobenius norm constraint as the decoder part to preserve the neighborhood structure.

\section{Sparse-induced Generative Adversarial Hashing (SiGAH)}\label{sec3}
\subsection{Overview}
In this section, we describe the proposed SiGAH in details.
Let $\mbX = \{\mbx_1,\mbx_2, ..., \mbx_n\}\in \mathbb{R}^{d\times n}$ be the dataset, which contains $n$ data samples each with a $d$-dimensional feature. 
For each data point $\mbx_i$, the hash function $\mathcal{H}(\mbx_i)$ produces an $r$-bit binary code $\mbb\in \{-1,1\}^r$ as:
\vspace{-0.5em}
\begin{equation} \label{eq2}
	\mbb = \mathcal{H}(\mbx_i) = \sign\big(\Func(\mbx_i)\big),
	\vspace{-0.2em}
\end{equation}
where $\sign(\cdot)$ is a sign function that returns $1$ if $\Func(\cdot) \geq 0$ and $-1$ otherwise.
Such a hash function is encoded as a mapping process combined with quantization.
$\Func(\cdot)$ is a linear or non-linear transformation function. 
In this paper, we mainly consider the linear hash function $\mcH(\mbx_i)=\sign(\mbW\mbx_i)$, where $\mbW\in \mbbR^{r\times b} $ is the projection matrix. 
Linear hash functions have been widely used in existing hashing works, including \cite{Gong2013IterativeQA,Heo2015SphericalHB,CarreiraPerpin2015HashingWB,Dai2017StochasticGH,Liu2017OrdinalCB,Datar2004LocalitysensitiveHS}, and can be easily extended to non-linear ones via the kernel tricks \cite{AAAI_liu,Jiang2015ScalableGH} or neural networks \cite{liong2015deep,Hu:2017uh,shen2019unsupervised}.

Inspired by \cite{balu2014beyond,Dai2017StochasticGH}, we consider a simple reconstruction as $\mbhx \propto \mbA \mcH(\mbx)$, where $\mbA \in \mbbR^{d\times r}$ is a reconstruction matrix.
For an input vector $\mbx$, the reconstruction based hashing targets at finding the best hash function, which involves minimizing the error of the feature reconstruction:
\vspace{-0.5em}
\begin{equation}\label{eq_2}
	Err(\mbhx,\mbx) = \|\mbhx - \mbx \|^2 = \|\mbA \mcH(\mbx) - \mbx\|^2.
	\vspace{-0.5em}
\end{equation} 
As shown in Eq.(\ref{eq_2}), finding the best hash function $\mcH$ is the key to minimizing the reconstruction error.
The proposed SiGAH framework achieves this by utilizing generative adversarial learning, which contains a generator and a discriminator.
In the following, we first briefly introduce the principles of the \emph{generator}, the \emph{discriminator}, and the \emph{hash learning}, the details of which are presented in Sec.\ref{sec31}, Sec.\ref{sec32} and Sec.\ref{sec33}, respectively.

\textbf{Generator.} We consider a generative model to generate a synthetic sample from a binary code, which ensures the distribution of the synthetic data is similar to that of the true data sample.
In order to obtain high-quality synthetic data, we first propose a \textit{sparsity-induced generative model} (SiGM).
SiGM first generates high-dimensional representations from the hash codes through an $L$-Lipschitz neural network, where the dimension of the generated feature is higher than that of original data. 
However, according to the quantized Johnson-Lindenstrauss lemma \cite{dasgupta1999elementary}, directly transforming the higher dimensional data to the original feature dimensions (\emph{e.g.}, by dimensional reduction via random projection)  will result in significant information loss. 
Therefore, we further introduce a sparsity constraint with a non-linear operation to the generative model, which contains a single hidden layer of Rectified Linear Unit (ReLU) activations \cite{ogcMdT}.
In the compressive sensing theory \cite{donoho2006compressed,jacques2015quantized}, such a  constraint ensures that the generated high-dimensional sparse vectors recover the original inputs effectively.

\textbf{Discriminator}. 
We further force the distribution of the recovered sample (synthetic data) to match the real data distribution. 
Our algorithm uses a discriminator to evaluate the quality of the synthetic data.
To that effect, the discriminator punishes the generator when it produces samples that are outside the real data distribution.
Following the recent work of \cite{zhao2016energy}, we use an energy function to construct our discriminator. 
This energy function is presented in the form of auto-encoders, where the reconstruction error serves as the energy scalar, which makes similar feature  distributions have lower energies.
We further use a marginal energy loss  to force the output of the generated model to be closer to the real data samples.

\textbf{Hash Learning}.
To train the hash functions, we replace the binary inputs of SiGM with the output of hash functions.
Consequently, a new generative neural network is constructed to generate synthetic data, which takes the original feature as input and the response of its latent layer as the output hash codes.
When the generator and the discriminator are well defined, the proposed generative adversarial framework can be used to train the hash functions.
For unsupervised hashing, it is usually necessary to closely encode data samples that have similar representations into the Hamming space.
We further add a graph regularization to the latent hash layer, which helps improve the quality of the hash codes.
The optimization can be performed by modifying the traditional Adam optimizer, which has shown to scale well to large datasets.
In the following, we introduce the three modules in detail.

\subsection{Sparsity-Induced Generative Model} \label{sec31}
Given a data point $\mbx\in \mbbR^d$, SiGAH aims to learn its corresponding hash code $\mbb\in \{0,1\}^r$.
To this end, we assume that there is a measurement $\mbPhi \in \mbbR^{r\times d}$ to approximate the reconstruction  as $\mbx = \mbPhi^T\mbb$.
As mentioned in \cite{balu2014beyond}, when the matrix $\mbPhi$ is orthogonal, the recovered feature has a lower reconstruction error.
Therefore,  we add an orthogonal constraint to matrix $\mbPhi$, and the relationship between $\mbx$ and $\mbb$ can be rewritten as:
\vspace{-0.5em}
\begin{equation} \label{eq3}
	\mbb = \mbPhi\mbx + \mbeta,~\textup{and}~ \mbx = \mbPhi^T(\mbb-\mbeta), ~\textup{where}~ \mbPhi^T\mbPhi = \mbI,
	\vspace{-0.5em}
\end{equation}
where $\mbeta\in \mbbR^{r}$ is the noise vector.
Although Eq.(\ref{eq3}) can be seen as a compressive sensing problem, $\mbx$ is always dense and $r \ll d$. 
Therefore Eq.(\ref{eq3}) is underdetermined, and cannot recover the original vector  $\mbx$ easily.

Following the theory of Restricted Isometry Property  \cite{candes2008restricted}, it is reasonable to assume a sparse generative function $\FG(\mbb)\!\!: \{0,1\}^r \rightarrow \mbbR^m$ that generates an $m$-dimensional sparse vector with a new measurement $\mbPsi\in \mbbR^{d\times m}~ (d \ll m)$, under a Gaussian distribution $\mcN(0,1/d)$, such that the original feature can be reconstructed via:
\begin{equation} \label{eq4}
	 \mbx = \mbPsi\FG(\mbb).
\end{equation}
Then, taking Eq.(\ref{eq4}) into Eq.(\ref{eq3}), a new recovery can be rewritten as follows:
\begin{equation}\label{eq5}
\mbb = \mbPhi\mbPsi\FG(\mbb) + \mbeta = \mbA\FG(\mbb) + \mbeta,
\end{equation}
where $\mbA = \mbPhi\mbPsi \in \mbbR^{r\times m}$.
That is, Eq.(\ref{eq5}) transforms the binary codes $\mbb$  to learn a sparse generative function $\FG(\mbb)$ under a new measurement $\mbA$. 
Therefore, the key issue is how to define the generative function.  
Due to the orthogonal constraint of $\mbPhi$,  $\mbA$ is still an $\mcN(0,1/d)$ Gaussian matrix.
Within such a context, the following proposition is shown as:

\textit{Theorem 1}\footnote{The corresponding proof is sketched in \cite{Bora2017CompressedSU}.}:  When the generative function $\FG(\cdot)$ is a $K$-layer neural network using ReLU activations, for any $\mby^*\!\in\! \mbbR^m$ and any observation $\mbb^*\! =\! \mbA\mby^*\!+\!\mbeta$, let ${\mbb}$ minimize $\|\mbb^*\! -\! \mbA\FG({\mbb})\|$ with an additive noise $\mbepsilon$ of the optimum.
Then, with $1\!-\!\exp^{-\mbOmega(r)}$ probability, we can achieve,
\begin{equation} \label{eq6}
	\|\FG({\mbb}) - \mby^*\| \leq 6 \min_{{\mbb}\in \mbbR^r}(\|\FG({\mbb})-\mby^*\| + 3\|\mbeta\| + 2\mbepsilon).
\end{equation}

However, due to the binary constraint of $\mbb \in \{-1,1\}^r$, we further extend such a boundary to the Hamming space, with a specific continuous relaxation process.
Let $\mc{B}_1=\{\mbbh_1,...,\mbbh_n| \mbbh_i\in [-1,1]^{r}\}$ and $\mc{B}_2=\{\mbz_1,...,\mbz_n | \mbz\in \{-1,1\}^{r}\}$, for all $\mbbh \in \mc{B}_1$, we have,
\begin{eqnarray}
	&\mc{D}^2(\mbbh, \mbz) \!=\! \min_{\mbbh} \sum_{k=1}^{r}(\mbbh_{k}-\mbz_{k})^2 \!= \!\sum_{k=1}^{r}(1-|\mbbh_{k}|)^2, \label{eq8}\\
	&\mc{D}(\mbbh, \mbz) =  \sqrt{\sum_{k=1}^{r}(1 - |\mbbh_{k}|)^2} \leq  \sum_{k=1}^{r}(1 - |\mbbh_{k}|)\nonumber  \\
	& \leq  \sum_{k=1}^{r}(1 - \mbbh_{k}^2) =  r  -  \Tr(\mbbh\mbbh^T ), \label{eq9}
\end{eqnarray}
where $\mc{D}$ is the Hamming distance  function. 
That is, the discrete binary codes in $\mc{B}_2$ can be approximated by the corresponding continuous set $\mc{B}_1$ with the upper bound constraint in Eq.(\ref{eq9}).
Then, a new boundary  can be given with the discrete constraint in Eq.(\ref{eq9}), which is presented as follows:
\begin{equation} \label{eq10}
	\|\FG({\mbbh}) - \mby^*\| \!\leq\! 6 \min_{{\mbbh}\in \mc{B}_1}\big(\|\FG({\mbbh})-\mby^*\| + \alpha \Tr({\mbbh}{\mbbh}^T) + 3\|\mbeta\| + 2\mbepsilon\big), 
\end{equation}
where $\alpha$ is the weight parameter, and $\FG({\mbb})$ can be seen as an optimal sparse approximation by giving the optimal hash codes $\mbb$.

Note that our goal is to find the optimal reconstruction of $\mbx^*$ with its input binary code $\mbb^*$.
To do this, we define the corresponding loss function as follows:
\begin{eqnarray} \label{eq7}
	\mcL(\FG,{\mbbh})& &=  \|\mbA\FG({\mbbh}) - \mbb^*\| \nonumber\\
	& &= \|\mbPhi\mbPsi\FG({\mbbh}) - \mbPhi\mbPhi^{T}\mbb^*\|, (\mbA = \mbPhi\mbPsi, \mbPhi\mbPhi^T = \mbI) \nonumber\\
	& &= \|\mbPhi\|\|\mbPsi\FG({\mbbh}) - \mbPhi^T\mbb^*\| =\|\mbPsi\FG({\mbbh}) - \mbPhi^T\mbb^*\|, 	\nonumber\\
	& &=\|\mbPsi\FG({\mbbh}) - \mbPhi^T\mbb^* + \mbPhi^T\mbeta - \mbPhi^T\mbeta \|, 	\nonumber\\
	& &\leq \|\mbPsi \FG({\mbbh}) - \mbPhi^T(\mbb^*-\mbeta)\| + \|\mbeta\|. 
\end{eqnarray}
Then, taking the inequation in Eq.(\ref{eq10}) into Eq.(\ref{eq7}), we can achieve the upper reconstruction boundary for any $\mbb^*= \mbA\mby^* + \mbeta$ and its corresponding $\mbx^*=\mbPsi\mby^*$ as:
\begin{align}
	&\frac{1}{6}\mcL(\FG, \mbbh) \nonumber\\
	& \leq \min_{\mbbh\in \mc{B}_1} \|\mbPsi\FG({
	\mbbh}) - \mbPhi^T\mbPhi\mbPsi\mby^*\| + \alpha\Tr(\mbbh\mbbh^T) + 4\|\mbeta\| + 2\mbepsilon , \nonumber\\
	& = \min_{\mbbh\in \mc{B}_1} \|\mbPsi\FG(\mbbh)-\mbPsi\mby^*\| + \alpha\Tr(\mbbh\mbbh^T) + 4\|\mbeta\| + 2\mbepsilon, \nonumber\\	
	&= \min_{\mbbh\in \mc{B}_1} \|\mbPsi\FG(\mbbh)-\mbx^*\| + \alpha\Tr(\mbbh\mbbh^T) + 4\|\mbeta\| + 2\mbepsilon. \label{eq11}
\end{align}
That is, when $\mbPsi$ is a random Gaussian matrix, recovering the original data via the generative function $\FG$ turns into minimizing the upper bound in Eq.(\ref{eq11}). 

Note that, as mentioned in \cite{Bora2017CompressedSU}, the  term $\mbepsilon$  from the gradient descent does not necessarily converge to the global optimum. Empirically, it typically converges to zero and can be ignored. 
Moreover, in order to ensure the sparsity of the output of $\FG$, we use a $K$-layer neural network with ReLU activations, which can automatically recover the support of the sparse codes \cite{ogcMdT}. 
Therefore, the loss function of the sparse-induced generative model (SiGM) with sparse regularization is defined as follows:
\begin{equation} \label{eq12}
\min\nolimits_{\mbb\in \mc{B}_1} \|\mbPsi\FG(\mbb)-\mbx\| + \alpha\Tr(\mbb\mbb^T) + 4\|\mbeta\| + \lambda \|\FG(\mbb)\|,
\end{equation}
where $\lambda$ is the weight parameter.
Moreover, the ReLU-based neural network (like the fully-connected layer) is the \emph{L}-Lipschitz, which has poly-bounded weights in each layer with $\mathcal{O}(rK\log m)$ sample complexity, where $K$ is the number of layers used in this neural network.
As shown in \cite{Bora2017CompressedSU},  a good input reconstruction can be obtained from the Gaussian matrix $\mbPsi$ using a ReLU-based generative function $\FG(\cdot)$.

\subsection{Discriminative Model} \label{sec32}
The key consideration of the discriminator lies in how to determine whether the recovered samples have a similar distribution to the original ones.
Due to the lack of label information, we redefine the original discriminator in the traditional GAN framework \cite{goodfellow2016deep} with an energy-based model, which maps each input to a single scalar, termed \emph{energy}.
As mentioned in \cite{zhao2016energy,lecun2006tutorial}, lower energy  can be  better attributed to the data manifold that reflects the real data distribution.

The generator can be viewed as a function that produces samples in a region of the feature space, to which the discriminator assigns higher energy.
Fortunately, an auto-encoder architecture is a good choice to define such an energy model, where the energy is the reconstruction error.
Therefore, the discriminator $\FD(\cdot)$ is structured as an auto-encoder:
\begin{equation} \label{eq13}
	\FD(\mba) = \big\|\Dec\big(\Enc(\mba)\big) - \mba\big\|,
\end{equation} 
where $\mba$ is the input data, and $\Dec$ \& $\Enc$ are the decoder function and encoder function, respectively.

The discriminator is trained to assign low energy to the regions in the feature space with high data density, and higher energy outside these regions. 
As a result, the discriminator with an energy function should give lower energy to the real data samples and higher energy to the generated ones.
Formally speaking, given the original data $\mbx$ and the synthetic data $\mbPsi\FG(\mbb)$, the discriminator loss can be defined as follows:
\begin{equation}
	\mcL_\FD(\mbx) = \mathbb{E}_{\mbx}\big[  \log \FD(\mbx)\big] + \mathbb{E}_{\mbb} \big[ \log\big( 1-\FD(\mbPsi \FG(\mbb))\big)\big].
\end{equation}
Similar to \cite{zhao2016energy},  a positive margin $\beta$ is further introduced into the discriminator in order to obtain better gradients when the generator is far from convergence.
\begin{eqnarray}
\mcL_\FD (\mbx,\mbb) = \mathbb{E}_{\mbx,\mbb}  \big[\FD(\mbx)+max\big[0, \beta - \FD\big(\mbPsi\FG(\mbb)\big)\big]\big]. \label{eq15}
\end{eqnarray}
As a result, a lower discriminator loss means the distribution of the synthetic samples matches the real ones better.

\subsection{Hash Learning}\label{sec33}

The final task is how to train hash functions $\mcH$ under the above generative adversarial framework.
Revisiting the proposed SiGM in Eq.(\ref{eq12}), the optimal generative function $\FG(\cdot)$ needs the optimal binary codes $\mbb$, obtained by minimizing the upper bound.
We further assume that the optimal binary codes can be generated from the hash functions in Eq.(\ref{eq2}).
However, the hash function is non-smooth and non-convex, which makes standard back-propagation infeasible for training such a generative model.

To solve this problem, we approximate the hash function using a one-layer fully-connected neural network with a hyperbolic tangent function $\delta(\cdot)$, which can be seen as a relaxed version of $\mcH(\mbx)\approx \delta(\mbW\mbx)$.
Therefore, a hash function can be performed to the input of generative function $\FG(\cdot)$, so that a new generative model $\hat{\FG}(\mbx) = \FG(\delta(\mbW\mbx))$ can be defined to approximate the high-dimensional sparse representation.

For unsupervised hashing, we also exploit the neighborhood structure in the feature space as an information source to steer the process of hash learning.
Inspired by \cite{Shaham:2018tz}, the in-batch similarity graph is introduced to the generative model as a regularization term.
Therefore,  the similarity between the $i$-th and $j$-th data is calculated by $\mbS(i,j) = \exp(-\sigma\|\mbx_i-\mbx_j\|)$, where $\sigma$ is a predefined hyper-parameter.
Similar to traditional graph hashing \cite{Liu2014DiscreteGH}, the objective loss fucntion for in-batch similarity can be written as:
\vspace{-0.5em}
\begin{equation}
	\mcL_S = \sum\nolimits_{i,j}^{N_B} \|\delta(\mbW\mbx_i) -\delta(\mbW\mbx_j) \| \mbS(i,j),
	\vspace{-0.5em}
\end{equation}
where $N_B$ is the number of data points in each training batch.

Furthermore, the signal-to-noise term $\|\mbeta\|$ can be further approximated with the quantization error $\|\delta(\mbx) - \mbW\mbx\|$.
We consider the hyperbolic tangent function with its first Talyor expansion as follows:
\vspace{-0.5em}
\begin{equation}
	\delta(\mbW\mbx) \!=\!   \delta(\mathbf{0}) + \frac{\delta^{'}(\mathbf{0})}{1!}\mbW\mbx + R_{n}(\mbW\mbx) \!=\! \mbW\mbx + R_{n}(\mbW\mbx),
	\vspace{-0.5em}
\end{equation} 
where $\delta^{'}$ is the first derivative of function $\delta$.
By substituting the above first-order Talyor expansion into the quantization error, we obtain the following equation:
\begin{align}
    \|\delta(\mbx) - \mbW\mbx\| &= \|\mbW\mbx - \mbW\mbx + R_{n}(\mbW\mbx)\| \nonumber\\ &=\|R_{n}(\mbW\mbx)\| \approx 0,
\end{align}
where $R_{n}(\mbW\mbx)\approx 0$.
As a result, the signal-to-noise term can also be ignored during training.

The final generative model with the graph regularization and discriminative compoments can be rewritten as:
\vspace{-0.5em}
\begin{align}
	\mcL_{\hat{\FG}}(\mbx) =& \mathbb{E}_{\mbx}\big[\|\mbPsi\hat{\FG}(\mbx) - \mbx\|  +  \alpha\Tr(\delta(\mbW\mbx)\delta(\mbW\mbx)^T) \nonumber\\
	&+ \lambda \|\hat{\FG}(\mbx)\|_2\big] + \gamma\mcL_S, \label{eq18} \\
	\mcL_\FD(\mbx) =& \mathbb{E}_{\mbx}  \big[\FD(\mbx)+max\big[0, \beta - \FD\big(\mbPsi\hat{\FG}(\mbx)\big)\big]\big], \label{eq19}
\end{align}
where $\alpha$, $\lambda$, $\beta$, and $\gamma$ are treated as hyper-parameters. 
Eq.(\ref{eq18}) and Eq.(\ref{eq19}) can still be considered a generative adversarial learning, which iteratively trains the discriminator and the generator to reach a Nash equilibrium.
The gradients of Eq.(\ref{eq18}) and Eq.(\ref{eq19})  can easily be obtained with respect to $\mbx$, and we can use SGD with the Adam optimizer to update the parameters \cite{Kingma2014AdamAM}.

\subsection{Anaylsis}
In this subsection, we analyze the connection between the proposed SiGAH and several existing algorithms.

\textbf{Connection to Binary Auto-Encoder (BA)} \cite{CarreiraPerpin2015HashingWB}:
If we replace the random matrix $\mathbf{\Psi}$ with an Identity matrix, delete the discriminative model in Eq.(\ref{eq19}) and prefix the parameters $\lambda$ and $\beta$ to zero, the optimization in Eq.(\ref{eq18}) can be reduced to Eq.(\ref{eq1}), which is the objective function of the binary auto-encoder.
Different from our method, a linear regression is used to reconstruct the original feature, which may achieve higher reconstruction error.
Moreover, BA requires significant resources and time to train the model.

\textbf{Connection to Stochastic Generative Hashing (SGH)} \cite{Dai2017StochasticGH}: SGH uses the $r$ dictionaries to reconstruct the input feature $x$.
When we reset the scale of the matrix $\mathbf{W}_2$ to $d\times r$,  delete the ReLU-layer, set the random matrix $\mathbf{\Psi}$ to the Identity matrix, and finally replace the sigmoid functions with its Doubly Stochastic Neuron, our proposed method will degenerate to the SGH method. 
Although SGH has achieved good performance on many ANN retrieval tasks, yet its reconstruction method is not the best: the scale of the dictionary is too small,  and  the original feature cannot be sufficiently recovered.

\subsection{The Importance of the Reconstruction} \label{sec30}
Given a data point $\mbx\in \mbbR^d$, SiGAH aims to learn the hash code $\mbb\in \{0,1\}^r$ and its corresponding encode function $\mbb = \mcH(x)$.
According to the loss function in Eq.(\ref{eq18}), SiGAH aims to minimize the error between $\mbx$ and $\hat{\mbx}$.
Therefore, we have the following error function:
\begin{eqnarray}
D_x(\hat{\mbx}_i,\hat{\mbx}_j) = D_x({\mbx}_i,{\mbx}_j) + Err,
\end{eqnarray}
where $Err$ is the reconstruction error.

Note that, $\mbb$ and $\mbx$ are in heterogeneous feature spaces, and the encode function $\mcH(x)$ has a  quasi-isometry property \cite{jacques2015quantized}.
Within such a context, the following proposition is shown as:

\textit{Proposition\footnote{The proof is similar to that in \cite{jacques2015quantized}.}:} 
A function $\mcH: x\rightarrow b$ is called a quasi-isometry between metric space $(\mbX, D_X)$ and $(\mbB, D_H)$ if there exists $C>0$ and $\Gamma \geq 0$:
\begin{equation}\label{eq30}
    \frac{D_X(\hat{\mbx}_i,\hat{\mbx}_j)}{C} - \Gamma \leq D_H(b_i,b_j) \leq C D_X(\hat{\mbx}_i,\hat{\mbx}_j) + \Gamma,
\end{equation}
where $x_i, x_j \in \mbX$, $b_i = \mcH(x_i)$, $D_X$ is the Euclidean distance, and $D_H$ is the Hamming distance.
As a conclusion, to reduce the reconstruction error, SiGAH should generate more robust binary codes $b_i$ and $b_j$.

\begin{figure}[!t]
\centering{
\includegraphics[width=0.5\linewidth]{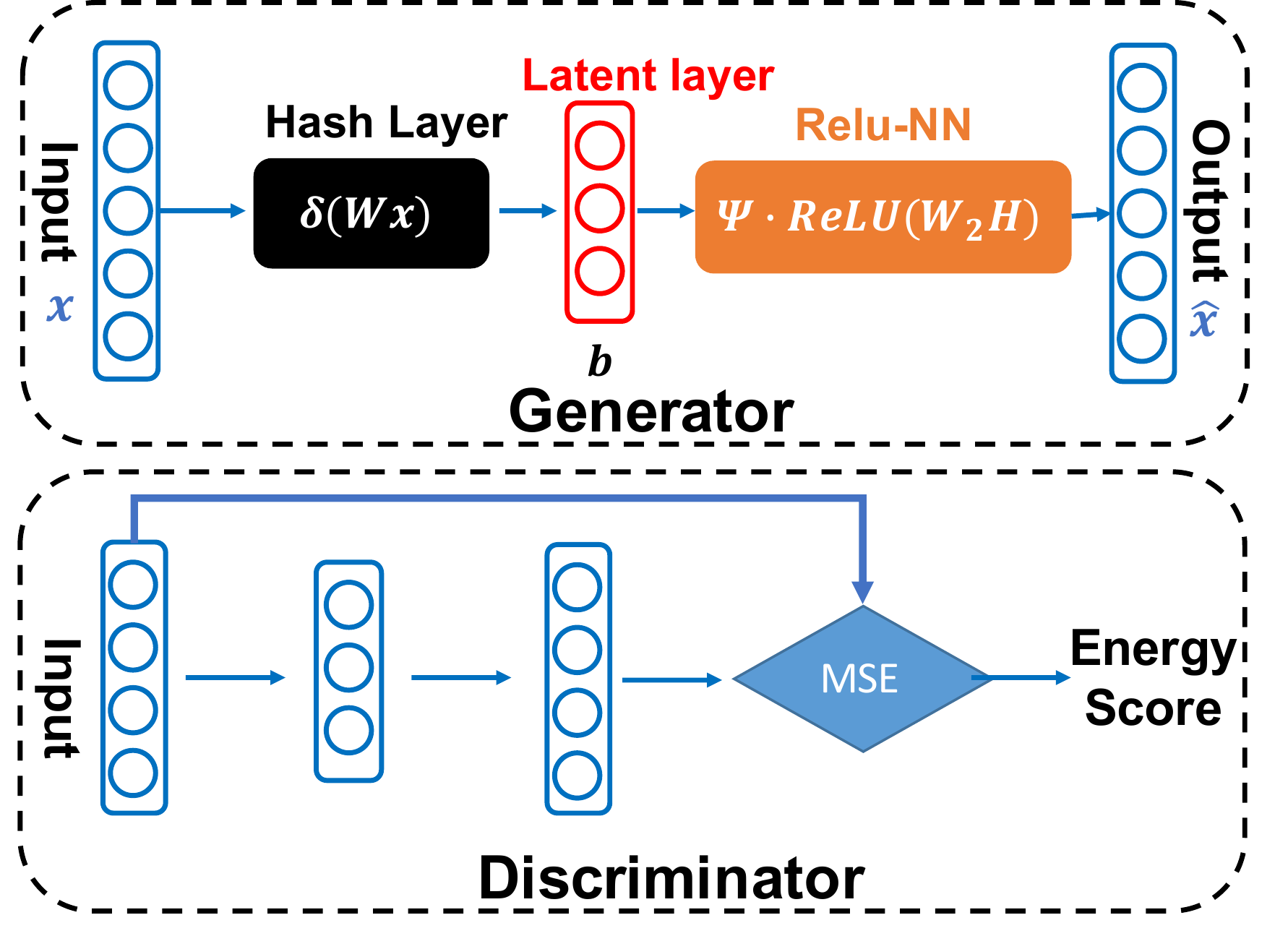}}
\caption{ Architecture setting of the SiGAH.\label{fig_stru}}
\end{figure}

\section{Implementation Details} \label{sec4}
We implement our SiGAH on the TensorFlow platform.
Before being rendered to the SiGAH networks, all the image features are extracted as mentioned in Sec.\ref{sec41}.
The proposed SiGAH contains two principle parts: a generator $\hat{\FG}$ and a discriminator $\FD$, whose structures are shown in Fig.\ref{fig_stru}.

\begin{table*}[!t]
\centering
\caption{The \emph{m}AP and Precision (Pre@100) comparison using Hamming Ranking on three benchmarks with different hash bits. (Boldface numbers indicate the best performance, while the second-best results are underline.)}
\label{tab1}\scalebox{0.7}[0.7]{
\begin{tabular}{c||c|c|c|c|c|c|c|c|c|c|c|c}
\hline
\multirow{2}{*}{method} & \multicolumn{4}{c|}{GIST1M}                                           & \multicolumn{4}{c|}{Deep1M}                                           & \multicolumn{4}{c}{Tiny100K}                                         \\ \cline{2-13} 
                        & \multicolumn{2}{c|}{mAP}          & \multicolumn{2}{c|}{Pre@100}      & \multicolumn{2}{c|}{mAP}          & \multicolumn{2}{c|}{Pre@100}      & \multicolumn{2}{c|}{mAP}          & \multicolumn{2}{c}{Pre@100}      \\ \hline
                        & 32 bits             & 64 bits              & 32 bits             & 64 bits             & 32 bits             & 64 bits              & 32 bits             & 64 bits              & 32 bits             & 64 bits              & 32 bits             & 64  bits            \\ \hline
LSH                     & 0.0897          & 0.1306          & 0.1841          & 0.2711          & 0.0875          & 0.1665          & 0.2359          & 0.4090          & 0.0596          & 0.1026          & 0.1434          & 0.2383          \\ 
ITQ                     & 0.1777          & 0.2090          & 0.3478          & 0.3976          & 0.1796          & 0.2926          & 0.4106          & 0.5924          & 0.1299          & 0.1558          & 0.2999          & 0.3504          \\ 
SpH                     & 0.1542          & 0.1972          & 0.3292          & 0.4150          & 0.1182          & 0.2183          & 0.3035          & 0.4993          & 0.1135          & 0.1680          & 0.2800          & 0.3993          \\ 
BA                      & 0.1715          & 0.1950          & 0.3426          & 0.3858          & 0.1919          & \underline{0.2975}    & 0.4266          & \underline{0.6090}    & 0.1114          & 0.1524          & 0.2749          & 0.3460          \\ 
SGH                     & \underline{0.1905}    & 0.2112          & \underline{0.3812}    & 0.4155          & \underline{0.1973}    & 0.2872          & \underline{0.4438}    & 0.5969          & \underline{0.1393}    & 0.1656          & \underline{0.3233}    & 0.3794          \\ 
OCH                     & 0.1675          & \underline{0.2355}    & 0.3409          & \underline{0.4437}    & 0.1754          & 0.2829          & 0.4121          & 0.5908          & 0.1266          & \underline{0.1893}    & 0.3009          & \underline{0.4088}    \\ 
\textbf{SiGAH}           & \textbf{0.2076} & \textbf{0.2594} & \textbf{0.3979} & \textbf{0.4837} & \textbf{0.2156} & \textbf{0.3307} & \textbf{0.4796} & \textbf{0.6510} & \textbf{0.1576} & \textbf{0.2425} & \textbf{0.3685} & \textbf{0.5046} \\ \hline
QoLSH ($d$-bit)             & -    & 0.2691    & -   & 0.4689          & -        & 0.5096          & -           & 0.8492         & -         & 0.1569       & -    & 0.3313 \\ \hline
\end{tabular}}
\end{table*}

For the generator, a good input reconstruction can be obtained from the Gaussian matrix $\mbPsi$ and ReLU-based neural network $\FG(\cdot)$.
Therefore, in this paper, we adopt the one-layer fully-connected  neural network with a LeakyReLU-based  activation as our generative function $\FG(\mbb)$, and we replace the input hash codes $\mbb$ by the hash function. 
Then, the hash function is defined as a one-layer fully-connected neural network and a hyperbolic tangent function is used as the activation function, which is added to the input of $\FG$. 
Specifically, the dimensionality of the latent layer is the given number of hash codes, and the parameter $m$ is always set to be twice the dimension of the input feature.

For the discriminator, we use a three-layer auto-encoder model with the nonlinear activation, the architecture of which is $[d-d_l-d]$.
We set $d_l=50$  for all experiments.
The parameter $\mbalpha$ is set to $0.1$ on Tiny100K, and $1e^{-3}$ on the other three datasets.
For the rest of the hyper-parameters, $\lambda$, $\beta$, and $\gamma$ are set to $1e^{-4}$, $0.1$, and $1e^{-4}$, respectively.
For all experiments, we use SGD to update the networks parameters, and the learning rate is set to $1\times 10^{-2}$ with a $5\times 10^{-4}$ weight decay.
The training batch size is fixed to $N_B = 500$, and all the weight parameters are initialized as random Gaussian matrices.

\section{Experiments} \label{sec5}
In this section, we evaluate our SiGAH scheme in comparisons to the state-of-the-art hashing methods \cite{Datar2004LocalitysensitiveHS,Gong2013IterativeQA,Liu2017OrdinalCB,CarreiraPerpin2015HashingWB,Dai2017StochasticGH} on four widely-used benchmarks, \emph{i.e.}, MNIST, Tiny100K, GIST1M, and Deep1M.

\subsection{Datasets} \label{sec41}
Four  large-scale image  retrieval benchmarks are used in this paper, which contain different image features:
\textbf{MNIST} \cite{lecun1990handwritten} contains $60,000$ digit images of size $28\times 28$ pixels. We use the flatted $784$-dimensional feature to represent  each image.
\textbf{Tiny-100K}  \cite{torralba2008small} contains a set of $1,00$K $384$-dimensional GIST descriptors, which are sampled from a subset of Tiny Images.
\textbf{GIST-1M} \cite{jegou2011product} is also widely-used to evaluate the quality of ANN search, and consists of one million GIST descriptors. 
\textbf{Deep1M} \cite{Babenko:2015uw} contains one million  $256$-dimensional deep descriptors, which are computed from the activations of AlexNet \cite{Krizhevsky2012ImageNetCW} with L2-normalization and PCA.

\begin{figure*}[!t]
\begin{center}
\begin{minipage}[t]{0.23\linewidth}
\centerline{
\subfigure[\footnotesize{R-$1$ on GIST1M}]{
\includegraphics[width=1.1\linewidth]{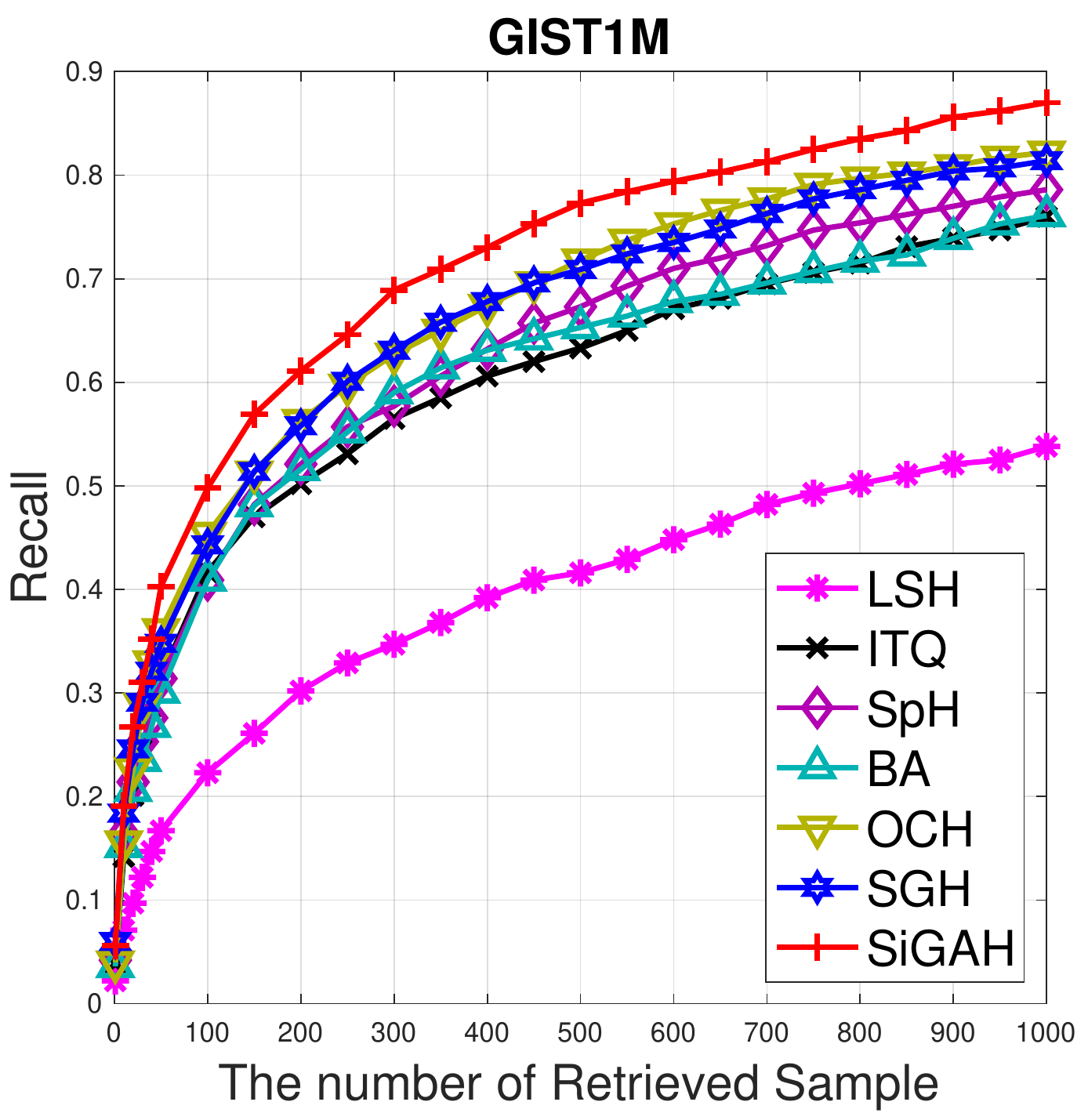}}\hspace*{-0.05\linewidth}
\subfigure[\footnotesize{R-$10$ on GIST1M}]{
\includegraphics[width=1.1\linewidth]{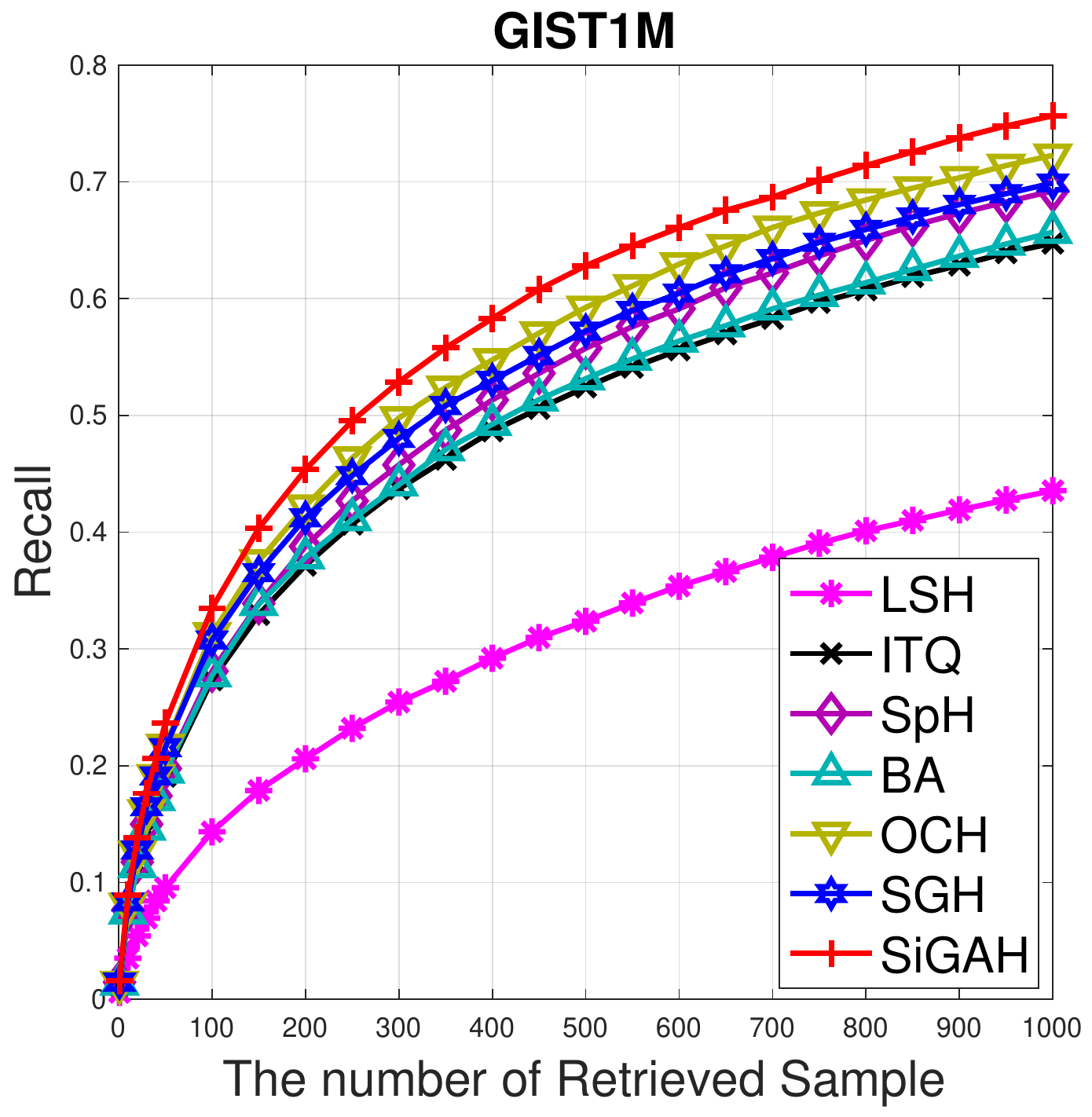}}\hspace*{-0.05\linewidth}
\subfigure[\footnotesize{R-$100$ on GIST1M}]{
\includegraphics[width=1.1\linewidth]{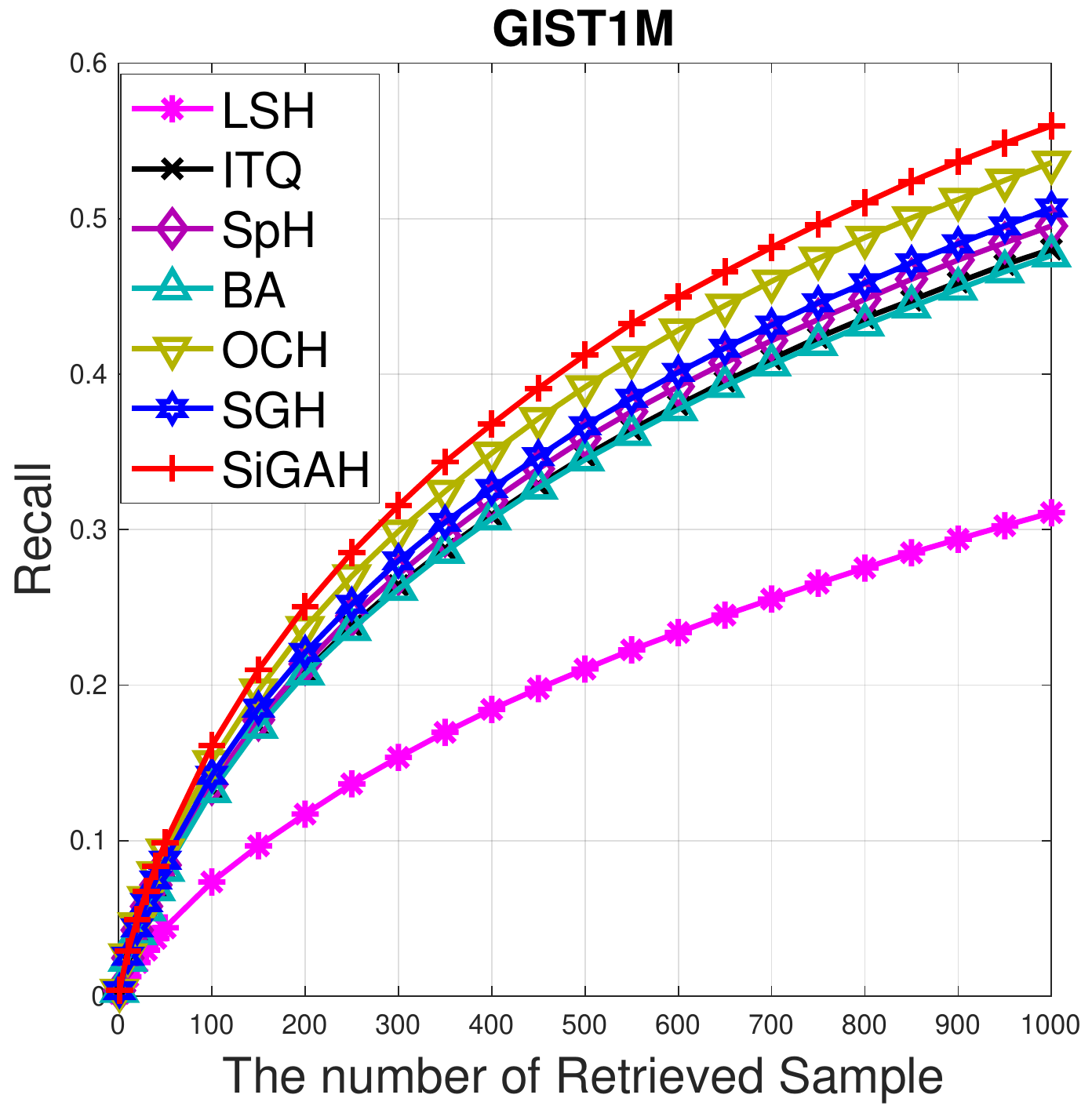}}\hspace*{-0.05\linewidth}
\subfigure[\footnotesize{R-$1$K on GIST1M}]{
\includegraphics[width=1.1\linewidth]{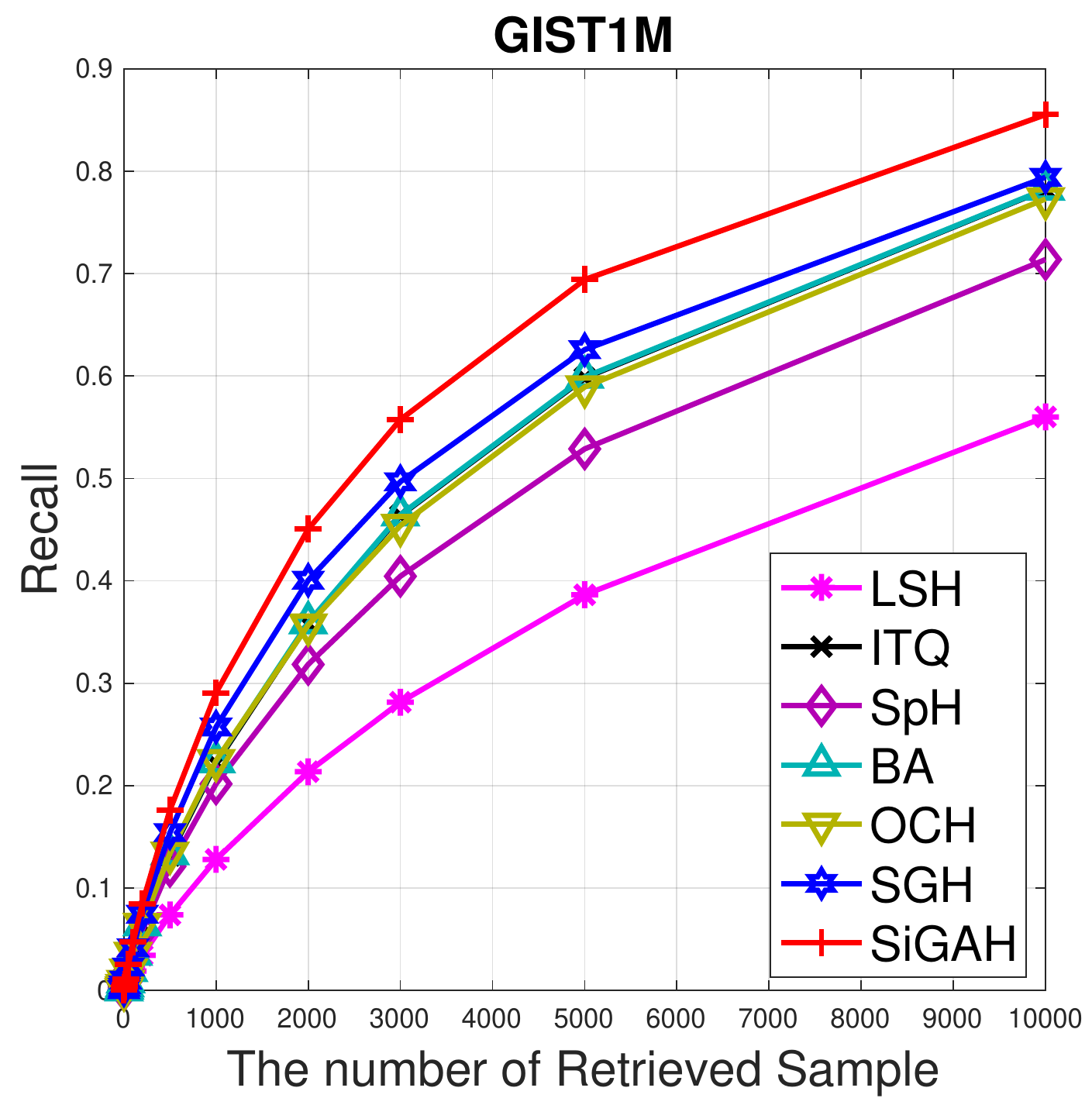}}
}
\end{minipage}

\begin{minipage}[t]{0.23\linewidth}
\centerline{
\subfigure[\footnotesize{R-$1$ on Deep1M}]{
\includegraphics[width=1.1\linewidth]{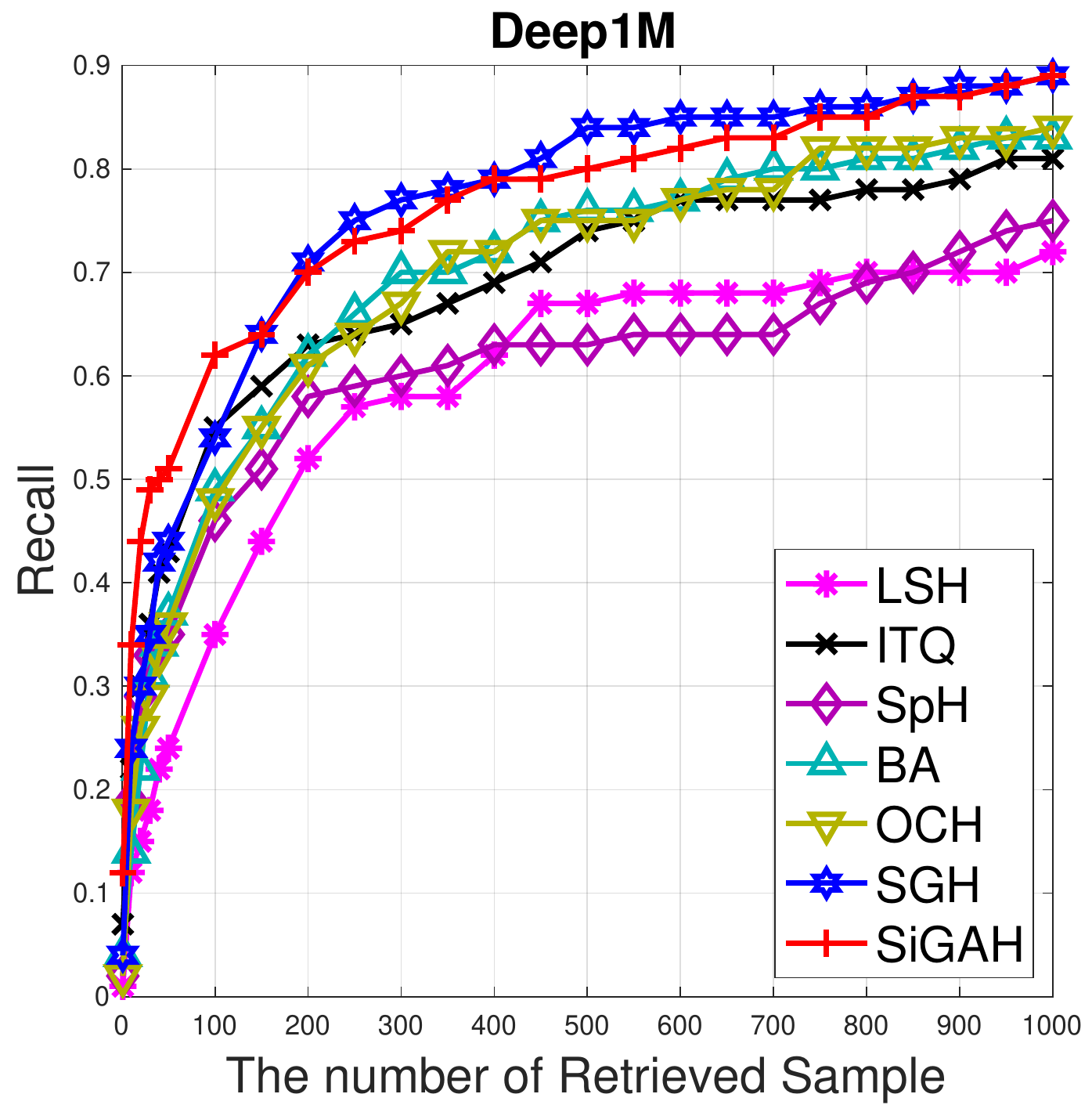}}\hspace*{-0.05\linewidth}
\subfigure[\footnotesize{R-$10$ on Deep1M}]{
\includegraphics[width=1.1\linewidth]{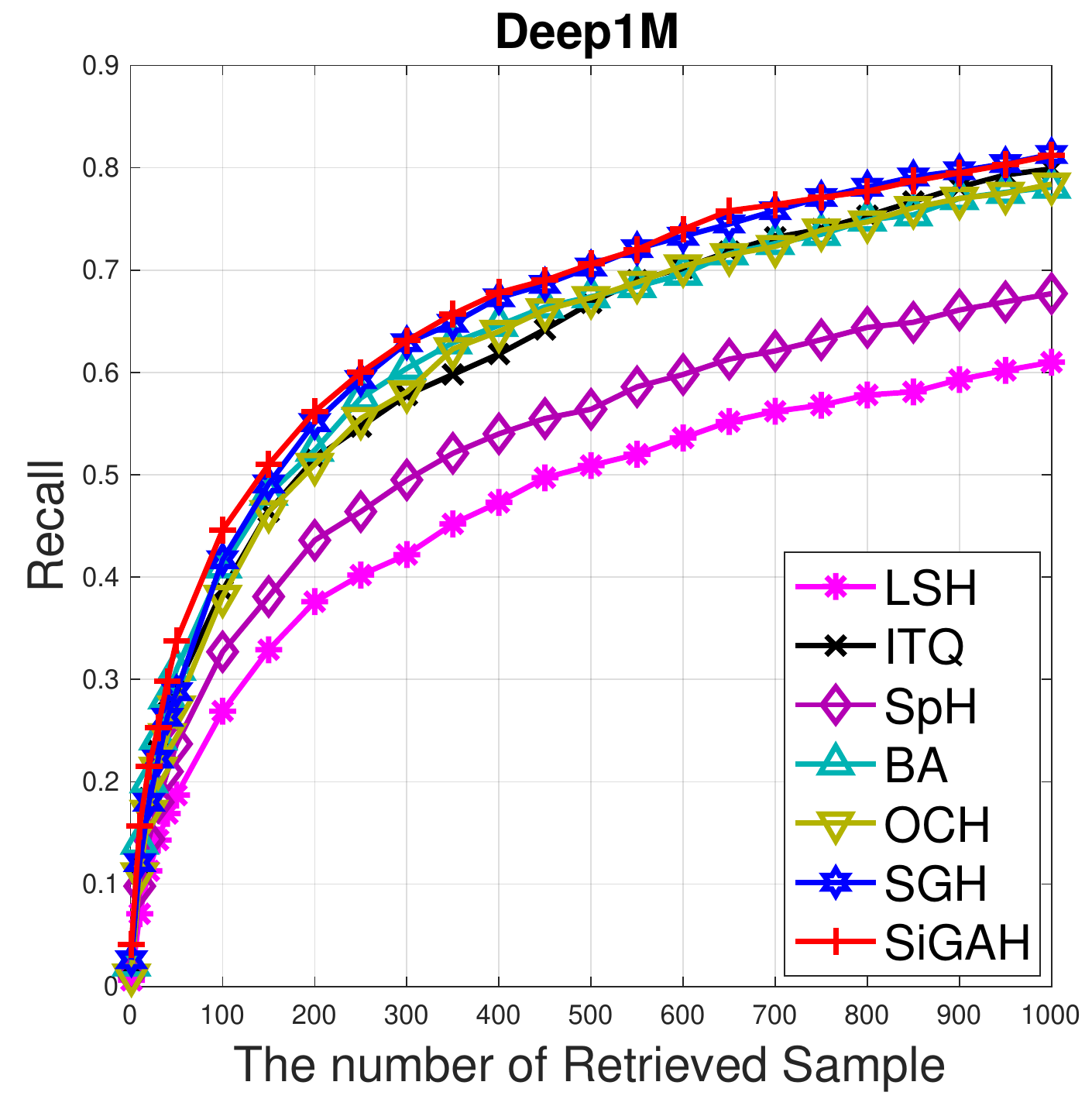}}\hspace*{-0.05\linewidth}
\subfigure[\footnotesize{R-$100$ on Deep1M}]{
\includegraphics[width=1.1\linewidth]{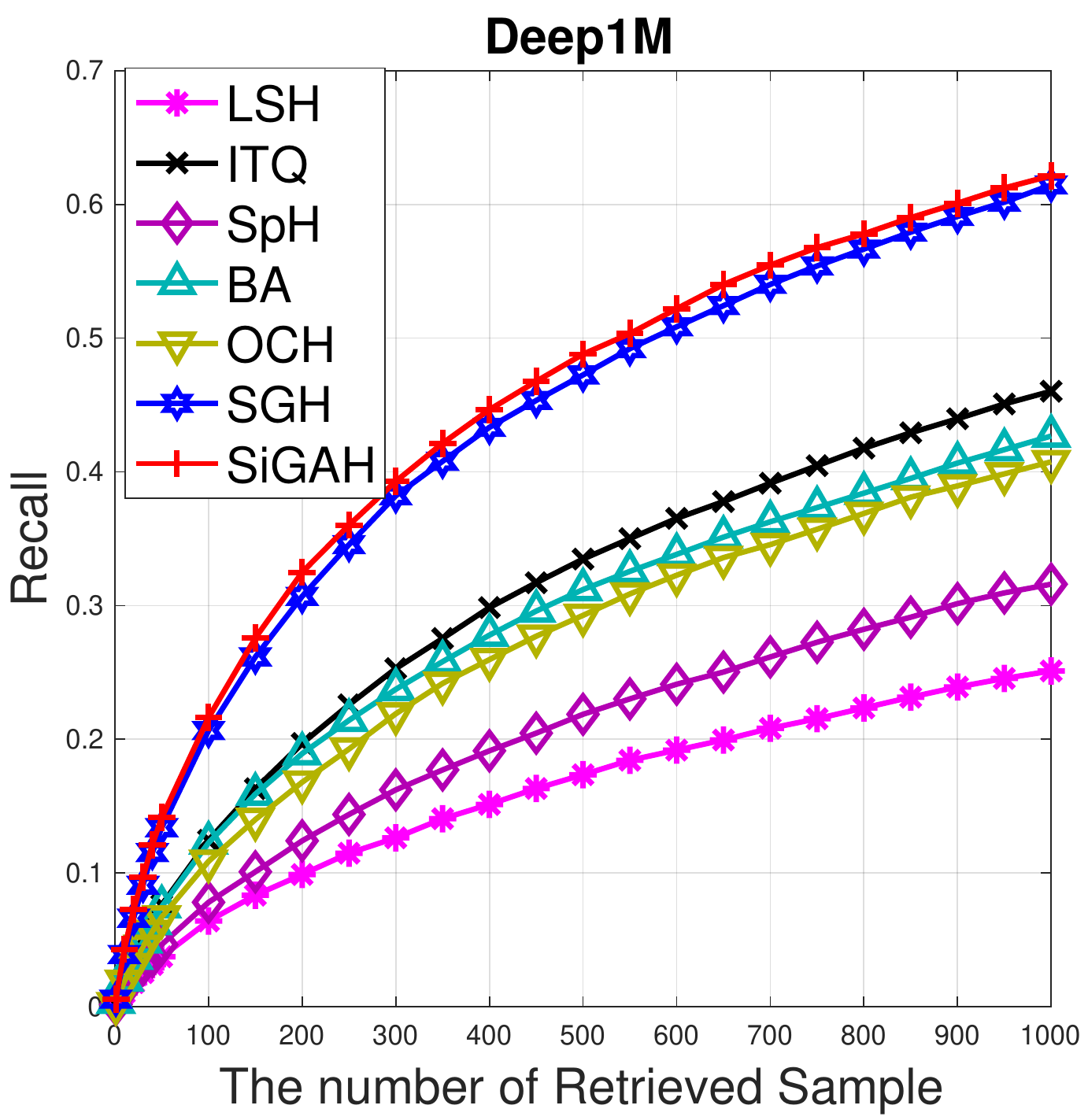}}\hspace*{-0.05\linewidth}
\subfigure[\footnotesize{R-$1$K on Deep1M}]{
\includegraphics[width=1.1\linewidth]{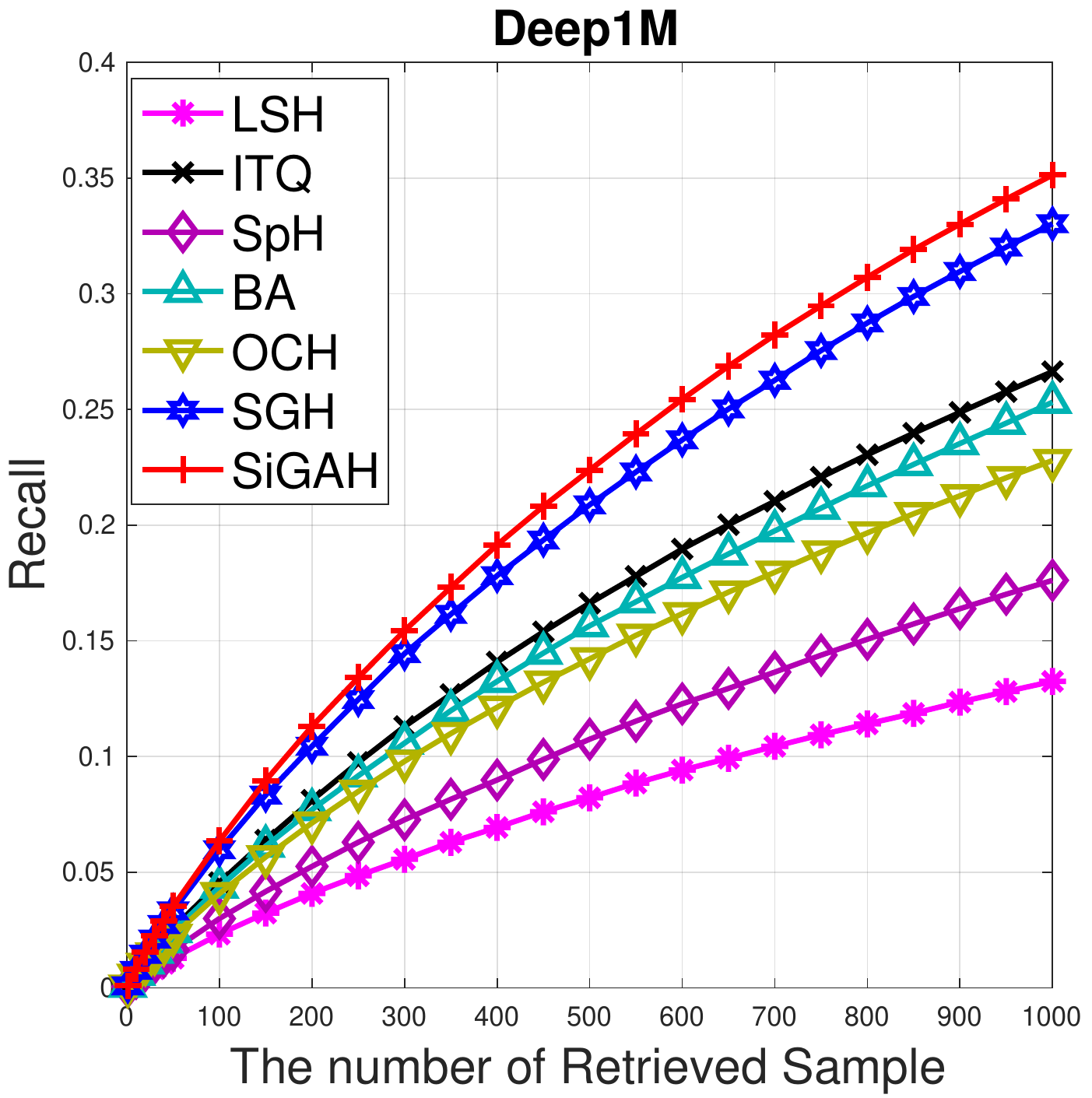}}
}
\end{minipage}

\begin{minipage}[t]{0.23\linewidth}
\centerline{
\subfigure[\footnotesize{R-$1$ on Tiny100K}]{
\includegraphics[width=1.1\linewidth]{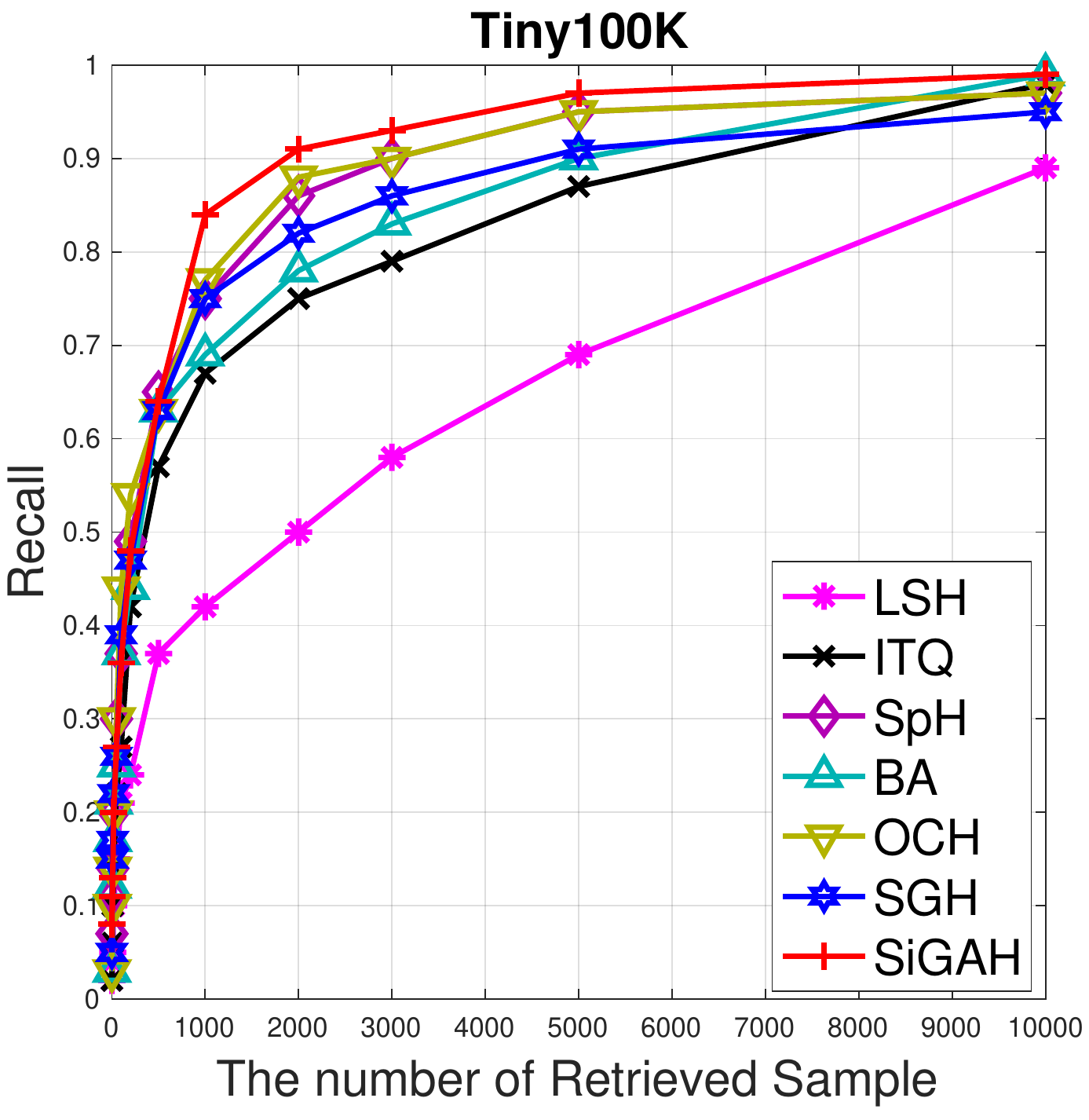}}\hspace*{-0.05\linewidth}
\subfigure[\footnotesize{R-$10$ on Tiny100K}]{
\includegraphics[width=1.1\linewidth]{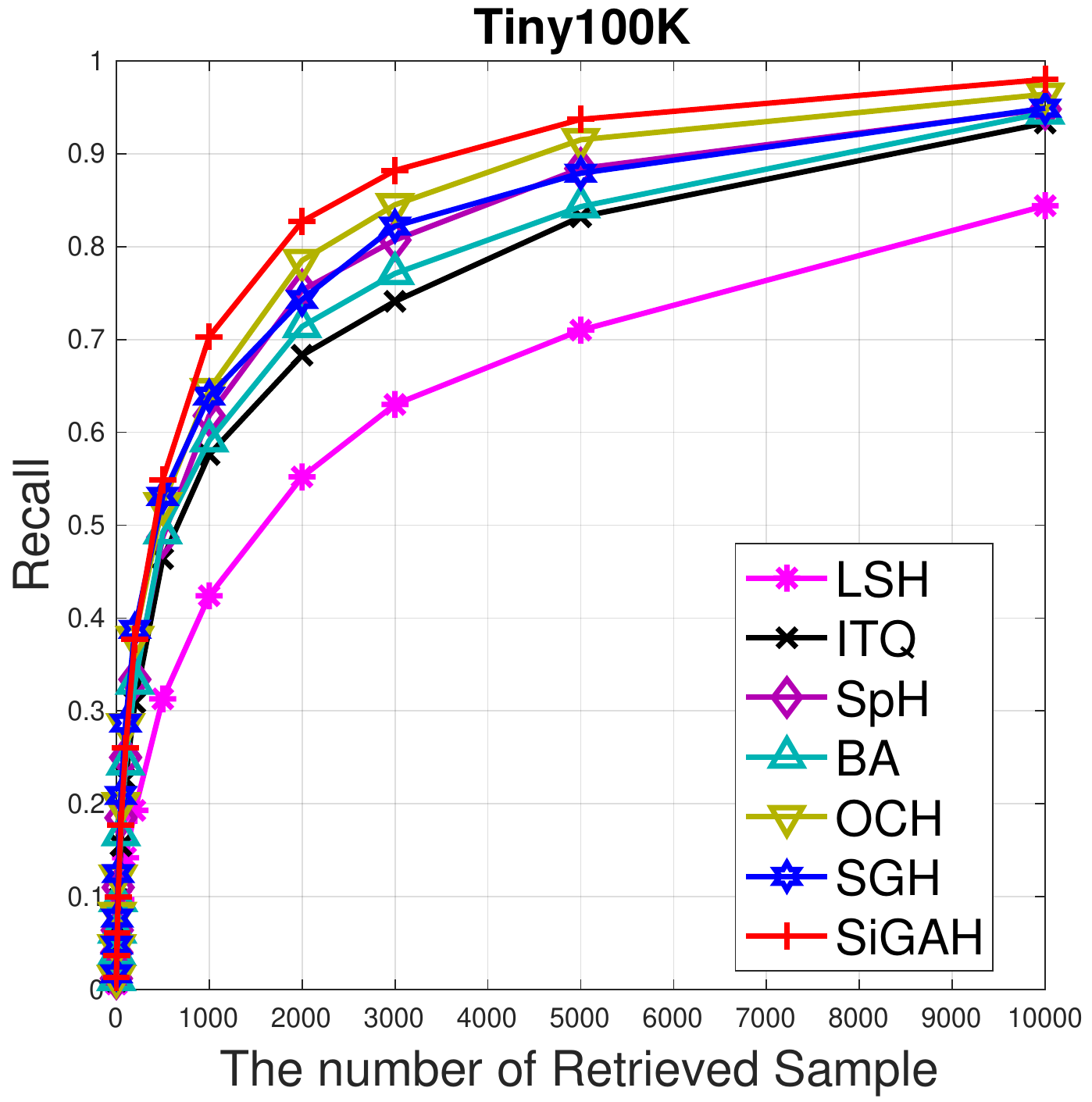}}\hspace*{-0.05\linewidth}
\subfigure[\footnotesize{R-$100$ on Tiny100K}]{
\includegraphics[width=1.1\linewidth]{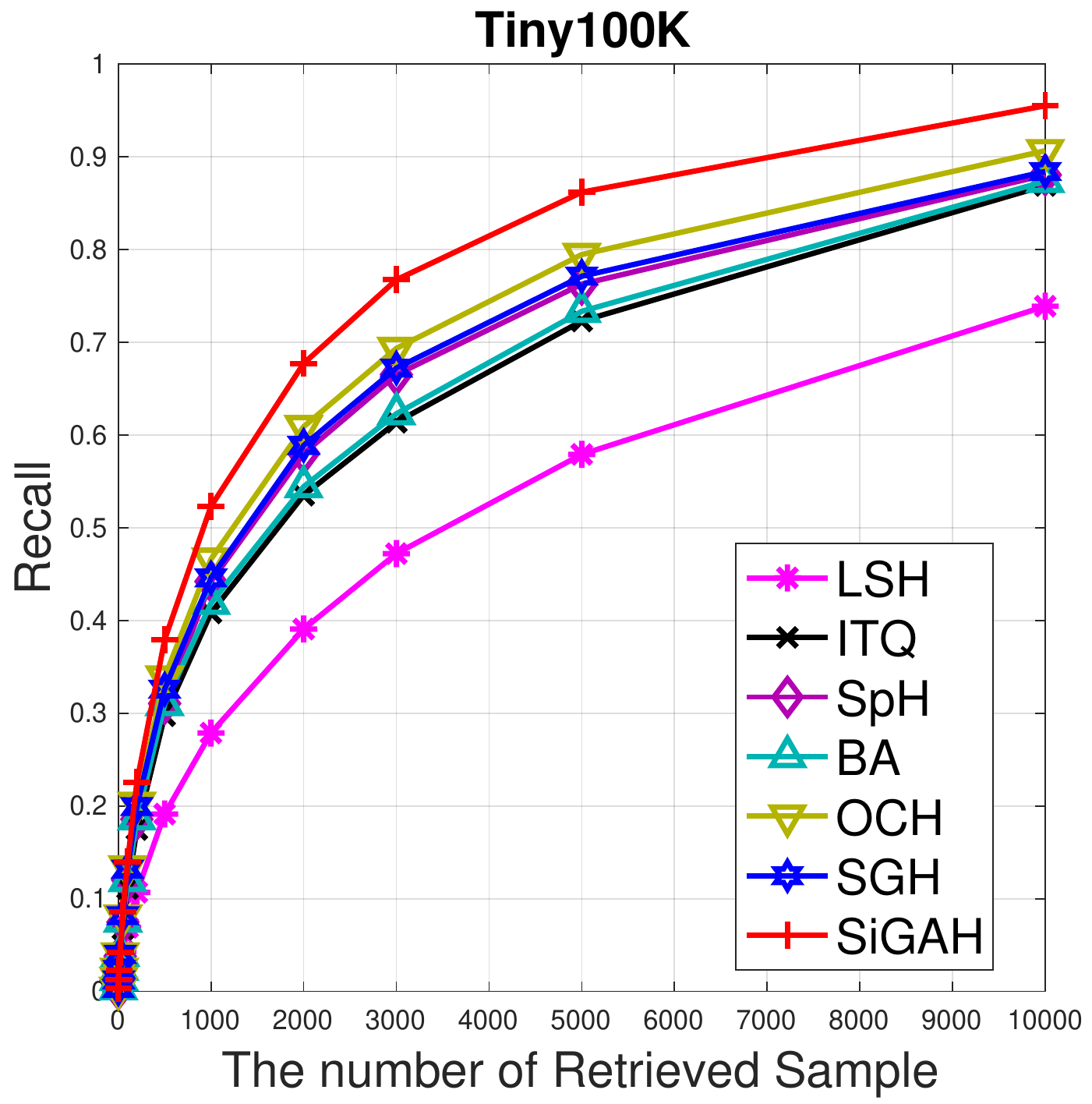}}\hspace*{-0.05\linewidth}
\subfigure[\footnotesize{R-$1$K on Tiny100K}]{
\includegraphics[width=1.1\linewidth]{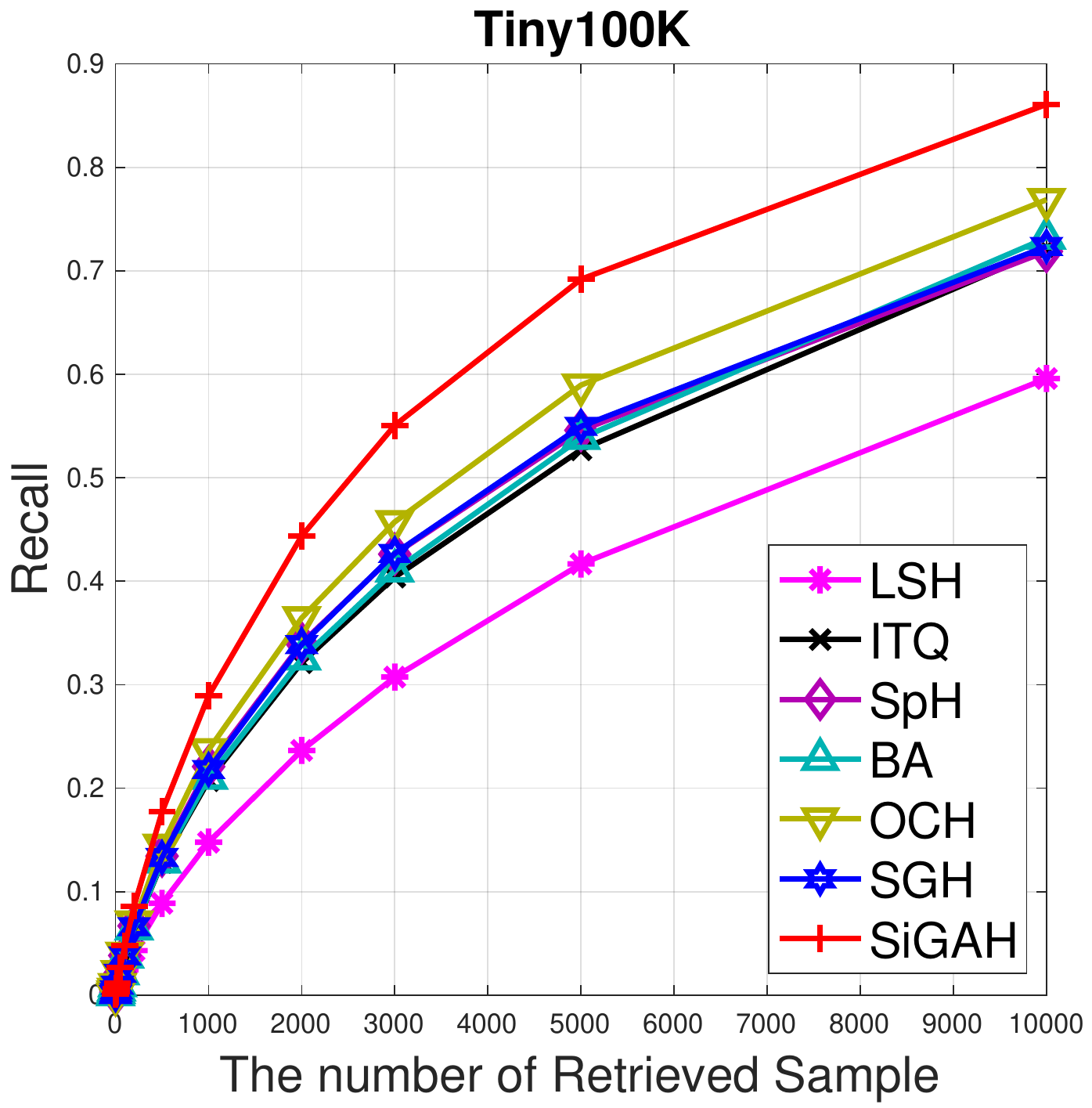}}
}
\end{minipage}
\end{center}
\caption{ANN search evaluation of different hashing methods on GIST1M, Deep1M, and Tiny100K with $64$ hash bits. (Best viewed in color.) \label{fig3}}
\end{figure*}

\subsection{Evaluation Protocols and Compared Methods} \label{sec42}
To evaluate the effectiveness of the proposed hashing algorithm, we follow the widely-used protocols in  \cite{Gong2013IterativeQA,Heo2015SphericalHB} to evaluate the ANN search, which mainly focuses on approximating the predefined metric.
Therefore, we use the top-$N$ ranking items with Euclidean neighbors as the ground-truth on these four datasets.
We set $N$ to $1,000$ as the default setting in most of our experiments. 
Then, based on the top-$1,000$ ground-truths, we compute the \emph{m}AP score, and Precision$@100$ (Pre$@100$).
However, using a smaller $N$ is generally more challenging, which better characterizes the retrieval performance. 
To this end, we further set $N$ as $1$, $10$, $100$ and $1,000$, and report the corresponding recall curves termed R-1, R-10, R-100, and R-1K, respectively.

\begin{figure*}[t]
\begin{center}
\begin{minipage}[t]{0.245\linewidth}
\centerline{
\hspace*{-0.05\linewidth}
\subfigure[Reconstruction vs. $\#$Iter.]{
\includegraphics[width=1.01\linewidth]{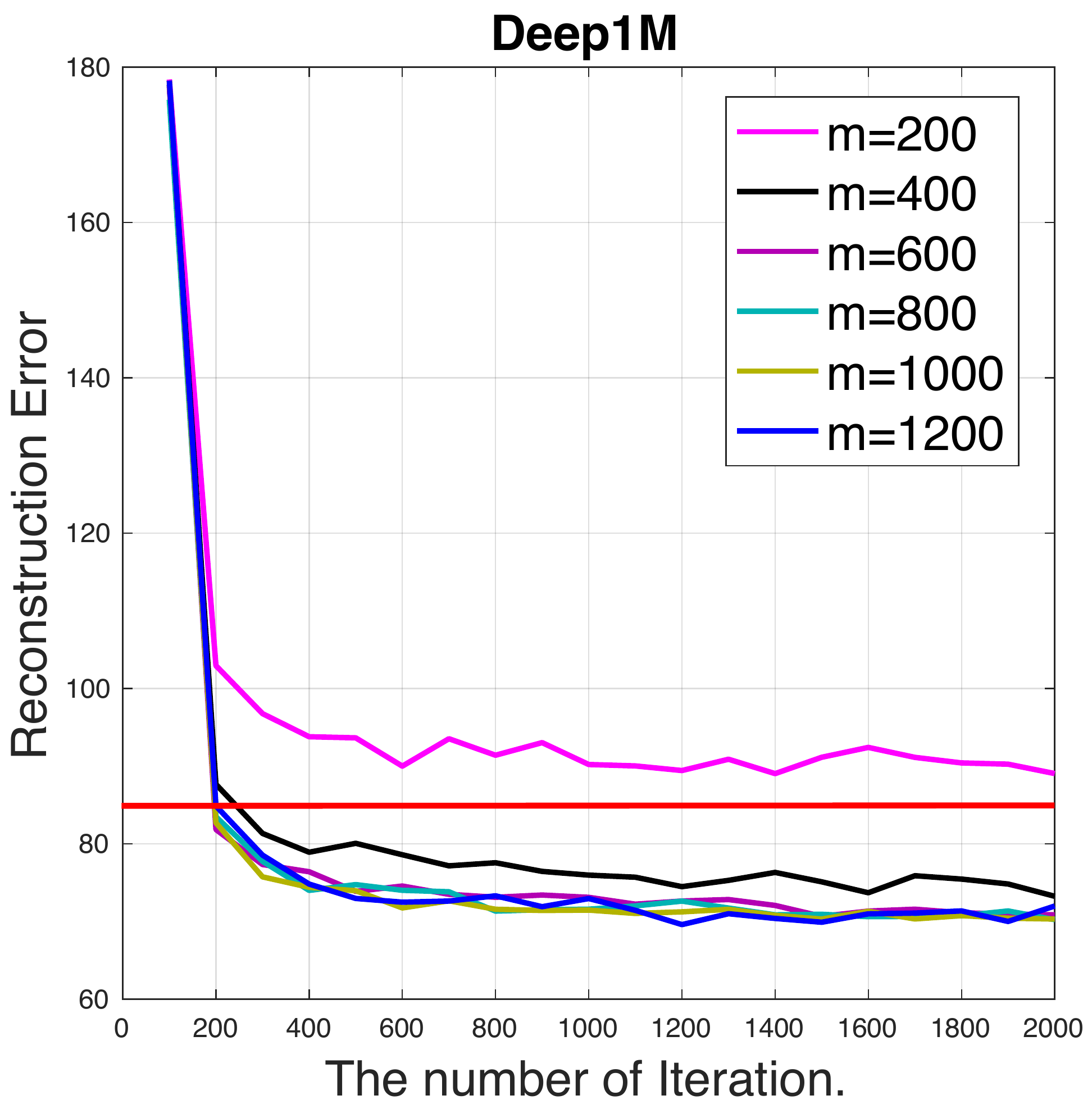}}\hspace*{-0.05\linewidth}
\subfigure[\emph{m}AP vs. $\#$Iter.]{
\includegraphics[width=1.04\linewidth]{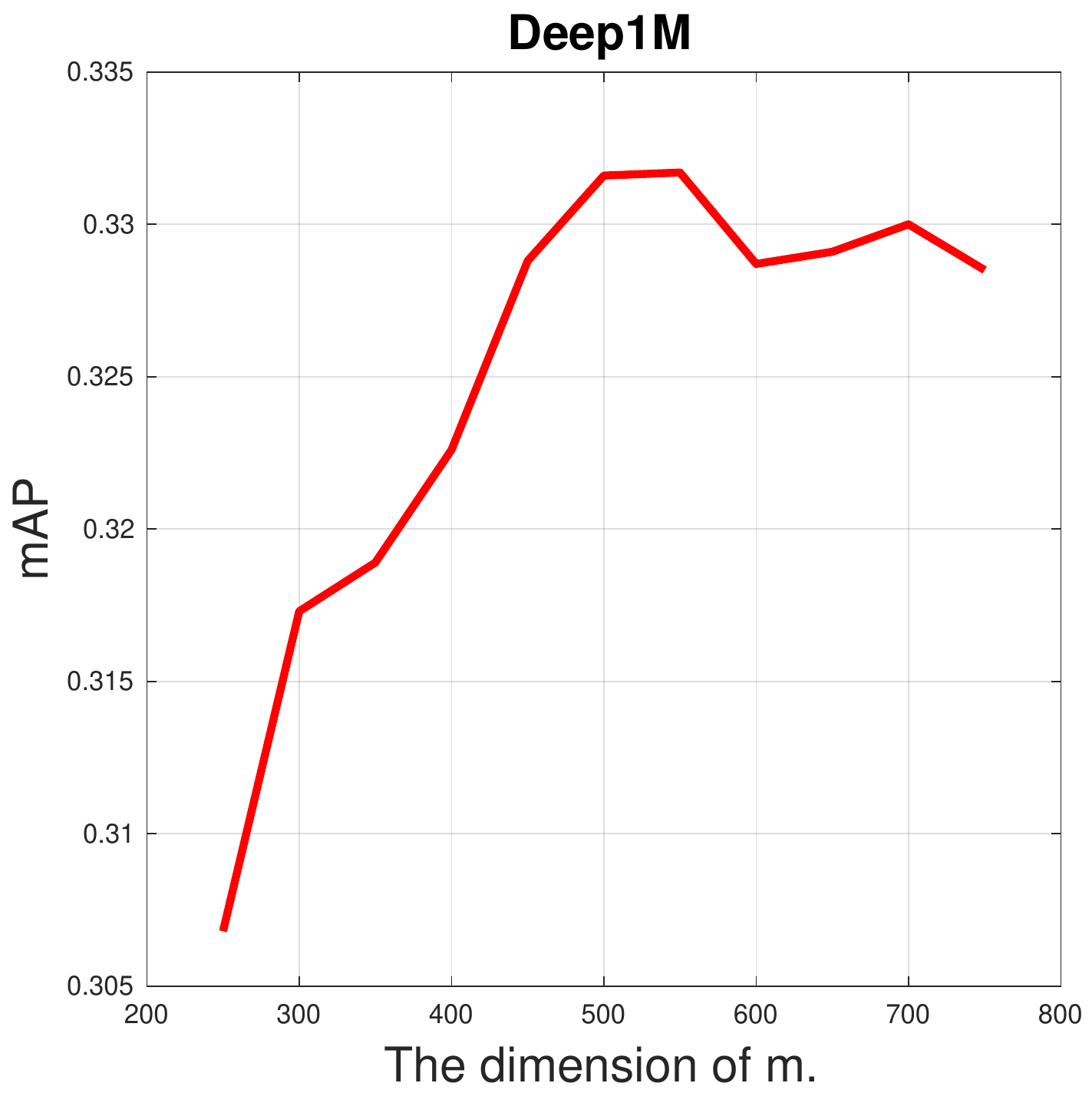}}\hspace*{-0.05\linewidth}
\subfigure[Reconstruction vs. $\#$Iter.]{
\includegraphics[width=1.01\linewidth]{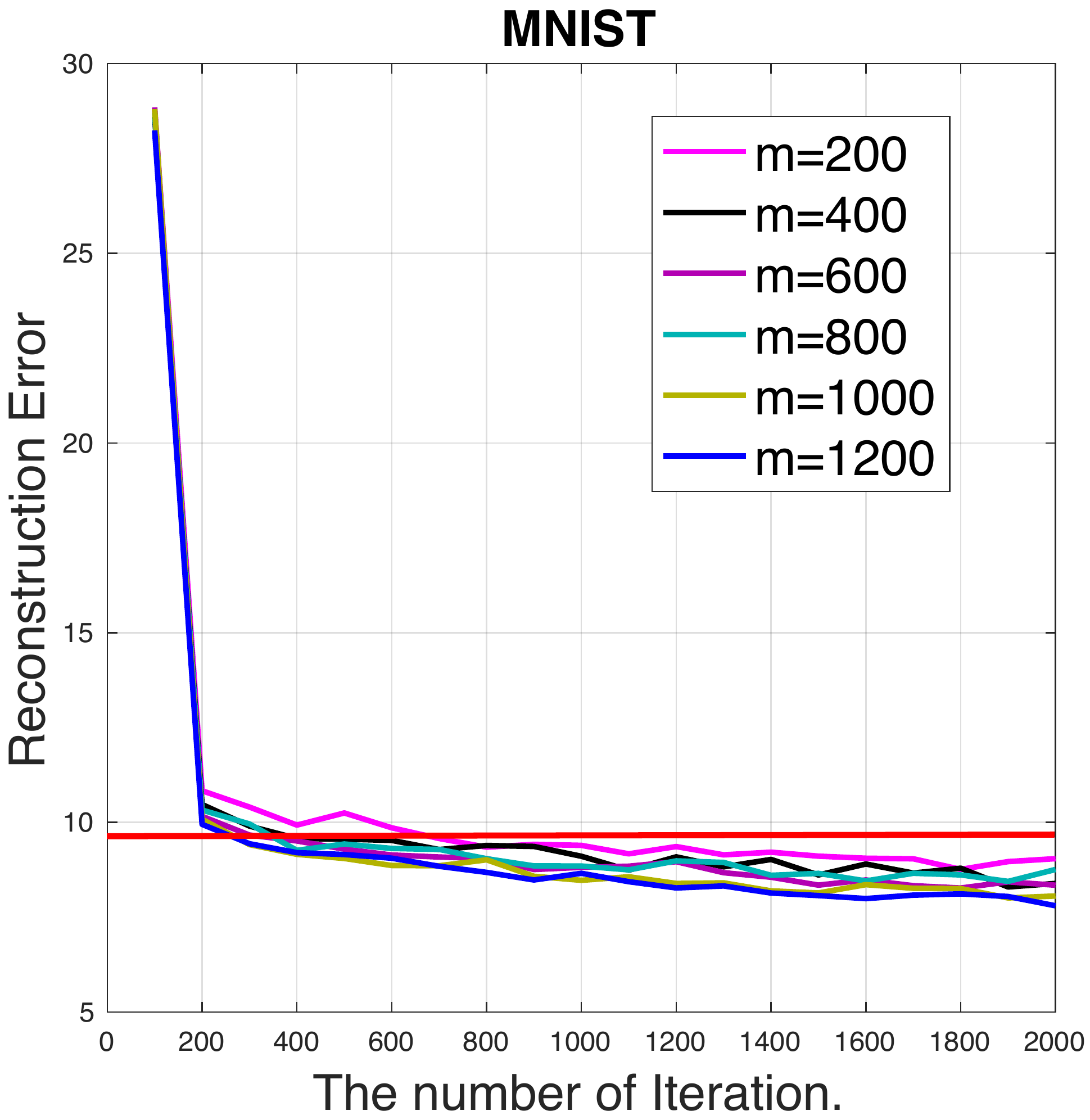}}\hspace*{-0.05\linewidth}
\subfigure[\emph{m}AP vs. $\#$Iter.]{
\includegraphics[width=1.04\linewidth]{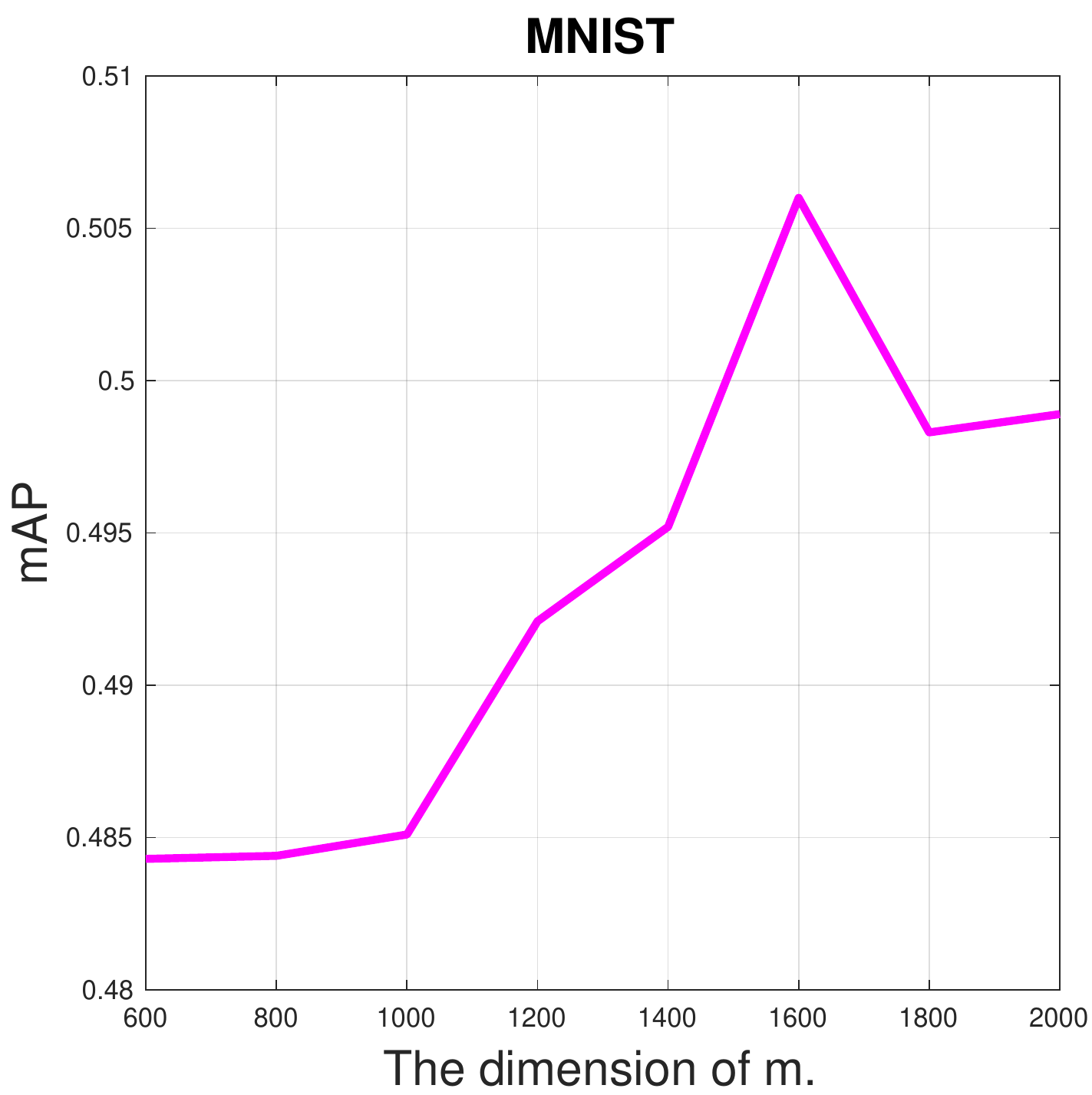}}
}
\end{minipage}
\end{center}
\caption{Reconstruction analysis of Deep1M and MNIST. Subfigure (a) and (c) present the relationship between  the reconstruction error and the scale of the random matrix $\mathbf{\Psi}$. Note that the red line is the reconstruction error of SGH. Subfigure (b) and (d) show the influence of the scale of the random matrix $\mathbf{\Psi}$. (Best viewed in color.)  }
\label{fig4}
\end{figure*}

\begin{figure}[t]
\begin{center}
\begin{minipage}[t]{0.42\linewidth}
\centerline{
\subfigure[Recall $@64$ on MNIST.]{
\includegraphics[width=0.6\linewidth]{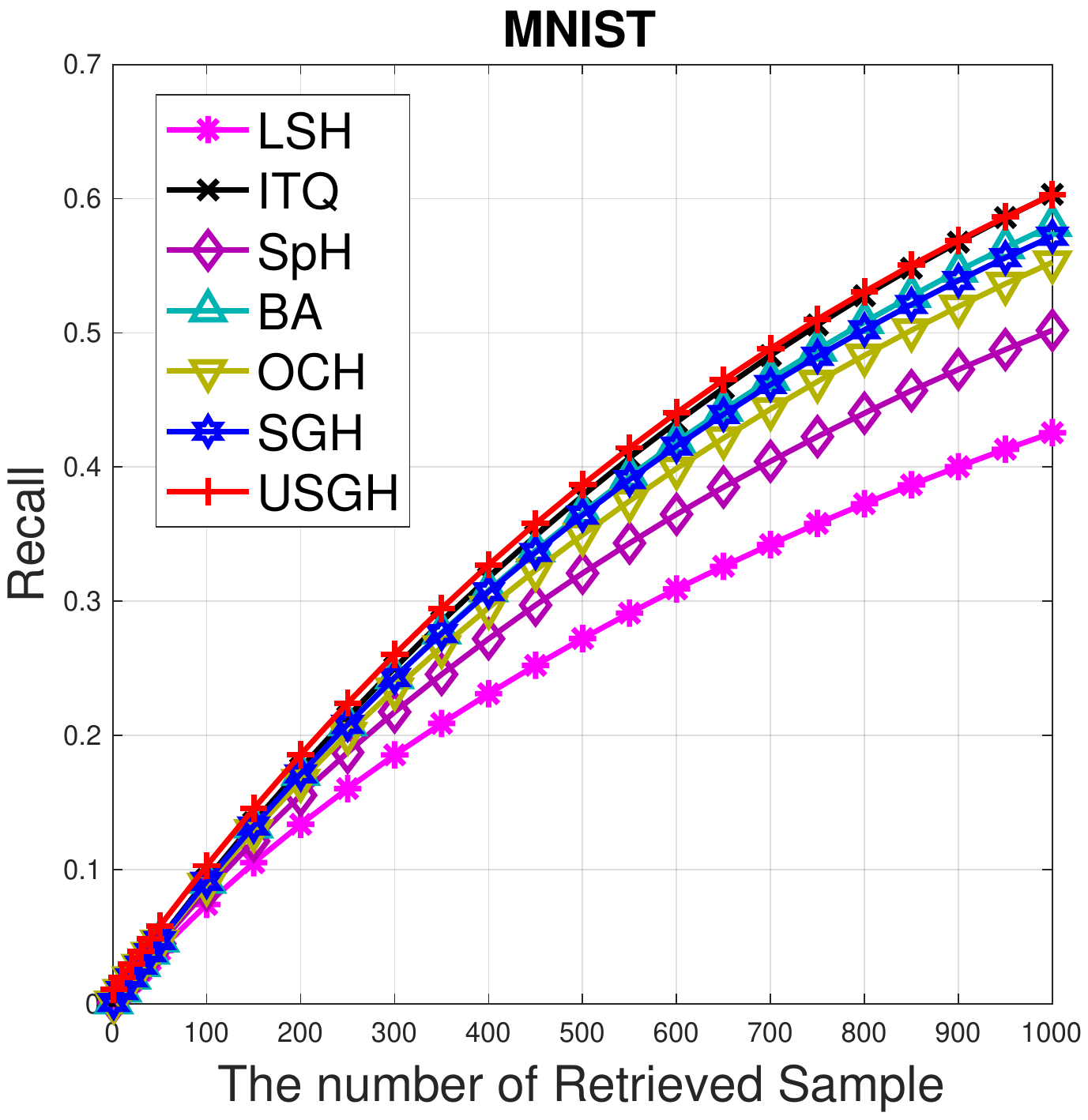}}
\subfigure[Precision $@64$ on MNIST.]{
\includegraphics[width=0.6\linewidth]{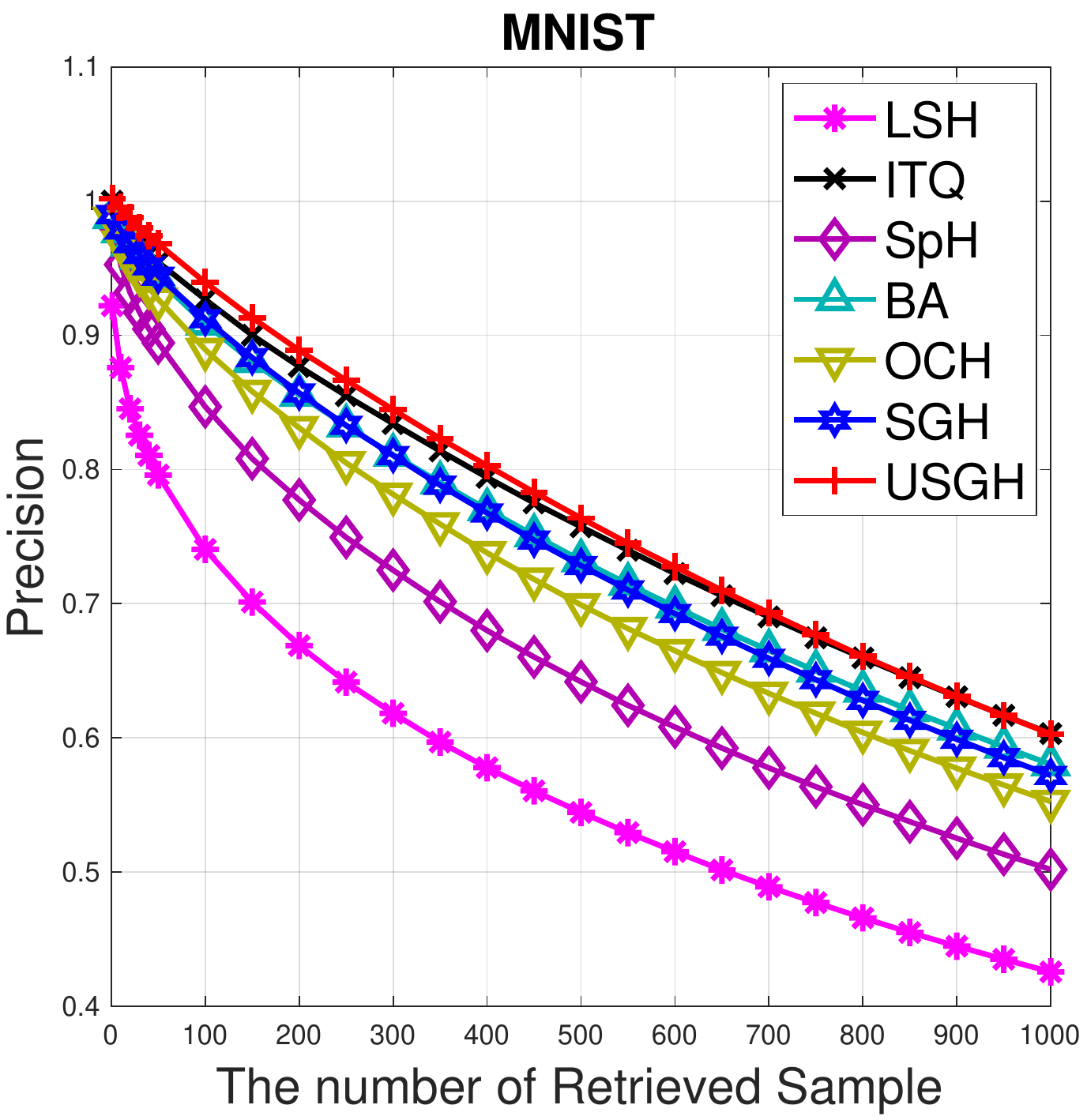}}
}
\end{minipage}
\end{center}
\setlength{\abovecaptionskip}{0pt}
\caption{Semantic search of performance of different hashing methods on MNIST  dataset. }
\label{fig5}
\end{figure}

We compare the proposed methods with six unsupervised hashing methods:
 Local Sensitive Hashing (LSH) \cite{Datar2004LocalitysensitiveHS}, Iterative Quantization (ITQ) \cite{Gong2013IterativeQA}, Spherical Hashing (SpH) \cite{Heo2015SphericalHB}, Binary Auto-encoder (BA) \cite{CarreiraPerpin2015HashingWB}, Stochastic Generative Hashing (SGH) \cite{Dai2017StochasticGH}, and Ordinal Constrained Hashing (OCH) \cite{liu2018ordinal}.
All the hashing methods above are unsupervised methods, including the proposed one.
The source codes of all the compared methods are kindly provided by the authors.
We implement our SiGAH hashing using TensorFlow, and we evaluate all the ANN search tasks on a single PC with Dual Core I7-3421 and 128G memory, where the complete dataset can be stored.\footnote{For fair comparison, all the training and testing are done on the CPU.}
We repeat all the experiments $10$ times and report the average performance over all runs.

\subsection{Quantitative Results} \label{sec43}

Table \ref{tab1} shows the experimental results under different coding lengths (\emph{i.e.}, 32 and 64). 
Obviously,  the proposed SiGAH method consistently achieves superior performance over all compared methods for all three datasets, \emph{i.e.}, GIST1M, Deep1M, and Tiny100K.
In fact, the average \emph{m}AP gain of SiGAH is more than $13.5\%$ across the  three datasets.
Furthermore, we find that the reconstruction-based hashing  (BA, SGH, and our scheme) can achieve much better results compared to other methods.
BA and SGH obtain the second best places, which demonstrates  that minimizing the reconstruction error is helpful for learning more accurate hash codes and hash functions.
Although OCH performs well with a coding length of  $64$-bit on GIST1M and Tiny100K, its performance using lower hash bits (such as $32$) is not satisfactory (and the proposed SiGAH performs much better than OCH with $19.16\%$ improvement).
In addition to the \emph{m}AP score, Tab.\ref{tab1} also shows the remarkable results for Pre@100, where SiGAH achieves an average improvement of $11.02\%$  compared to the second best hashing algorithm (the underline results \emph{i.e.}, SGH, BA and OCH).
Note that we also evaluate the reconstruction-based method QoLSH \cite{balu2014beyond}.
Although the overall performance of QoLSH is better on these three datasets, QoLSH  requires that the hash bit be larger than the dimension of input feature (set to the dimension of input data in our experiment), making it less practical and less flexible compared to our SiGAH.
Apart from on the Deep1M dataset, SiGAH consistently achieves better results  than QoLSH, which demonstrates that the proposed generative model is  advantageous for producing more distinguished codes.


Following the experimental setting of \cite{he2013k}, we further plot the recall curves  for the GIST1M, Tiny100K, and Deep1M datasets when the hash bit is set to $64$. 
The results are shown in Fig.\ref{fig3}.
The recall is defined as the fraction of retrieved true nearest neighbors to the total number of true nearest neighbors.
Recall-$N$ (R-$N$) is the recall of $N$ ground-truth neighbors in the top retrieved samples.
Note that a smaller $N$ is generally a more challenging criterion, which better characterizes the retrieval results. 
Fig.\ref{fig3} shows that the overall performance of our SiGAH is best across all  three datasets, where all algorithms use the same search scheme with the same search time.
\begin{table}[]
\centering
\caption{\emph{m}AP and Precision using Hamming ranking on MNIST for the semantic retrieval task. (Boldface is the best performance, and the second-best results are underline.)}
\label{tab2}
\scalebox{1.0}[1.0]{
\begin{tabular}{c||c c|c c}
\hline
\multirow{2}{*}{Method} & \multicolumn{4}{c}{MNIST}                                                                            \\ \cline{2-5} 
                        & \multicolumn{2}{c|}{mAP}          & \multicolumn{2}{c}{Pre@100}       \\ \hline
                        & 32              & 64              & 32              & 64                       \\ \hline
LSH                 & 0.2148       & 0.3598       & 0.4952       & 0.7450                \\ 
ITQ                  & \underline{0.4223}          & 0.4681       & \underline{0.8255}         & \underline{0.9267}               \\ 
SpH                 & 0.3462          & 0.4894          & 0.6786          & 0.8467                  \\ 
BA                      & 0.3897          & 0.4156          & 0.8045          & 0.9086        \\ 
SGH                     & 0.4195          & \underline{0.4920}          & 0.8101          & 0.9124         \\ 
OCH                     & 0.4191          & 0.4550          & 0.7649          & 0.8835               \\ 
\textbf{SiGAH}           & \textbf{0.4320} & \textbf{0.5014} & \textbf{0.8255} & \textbf{0.9294}  \\ \hline
\end{tabular}}
\end{table}

Finally, we also evaluate the  retrieval performance using MNIST.
The results for the Pre@100 and \emph{m}AP are reported in Table \ref{tab2}.
SiGAH still obtains the best retrieval performance compared to all other methods. 
In terms of precision score, SiGAH achieves a $2.71\%$ gain over the second best baseline, \emph{i.e.}, ITQ \cite{Gong2013IterativeQA}.
Note that ITQ gets second best place mostly in this retrieval task.
To explain, ITQ can also be viewed a reconstruction-based hashing \cite{Wang:2017gy}. 
As mentioned in Section \ref{sec31}, the objective of ITQ can be considered a modification of Eq.(\ref{eq5}), where the original features are directly reconstructed from the binary codes with the orthogonal constraints.

\begin{figure}[!t]
\centering{
\includegraphics[height=0.35\linewidth]{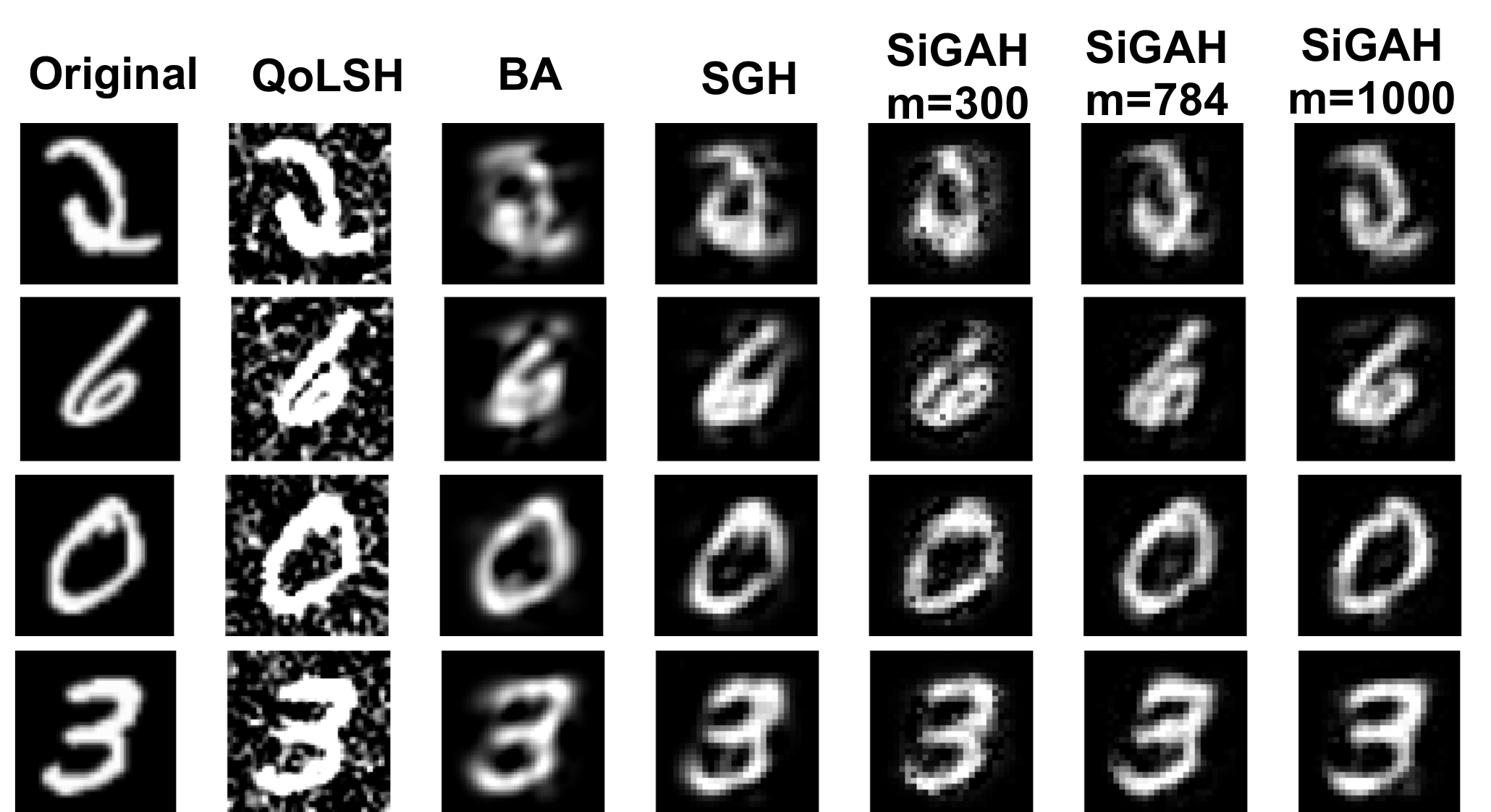}}
\caption{Reconstructed images on MNIST with BA, SGH, and SiGAH on $64$-bit. (QoLSH is on $784$-bit.)\label{fig_gen}}
\end{figure}

\begin{figure*}[t]
\begin{center}
\begin{minipage}[t]{0.23\linewidth}
\centerline{
\hspace*{-0.05\linewidth}
\subfigure[\emph{m}AP vs.  $\beta$.]{
\includegraphics[width=1.1\linewidth]{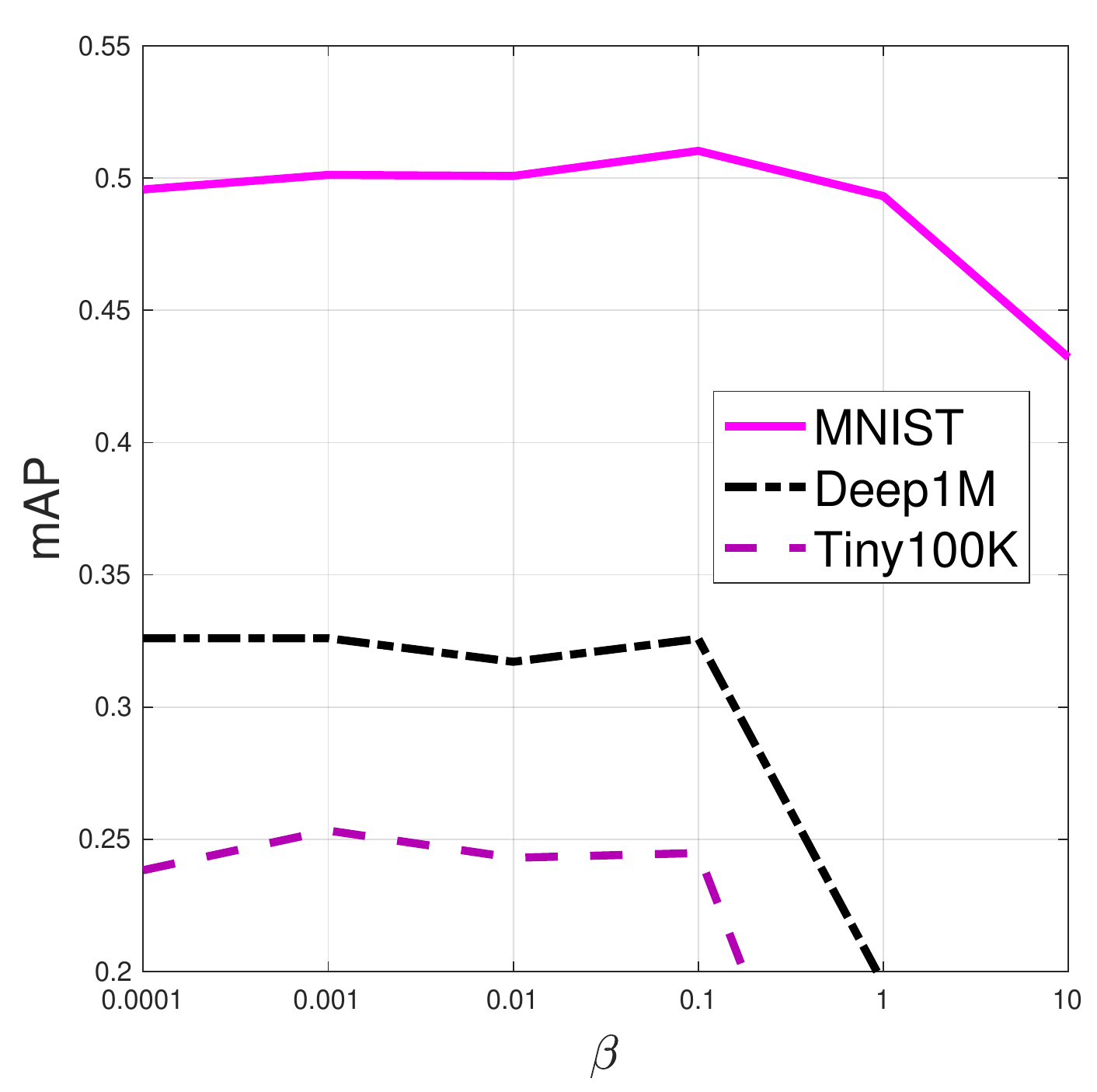}}\hspace*{-0.05\linewidth}
\subfigure[Energy vs. $\#$Iter.]{
\includegraphics[width=1.09\linewidth]{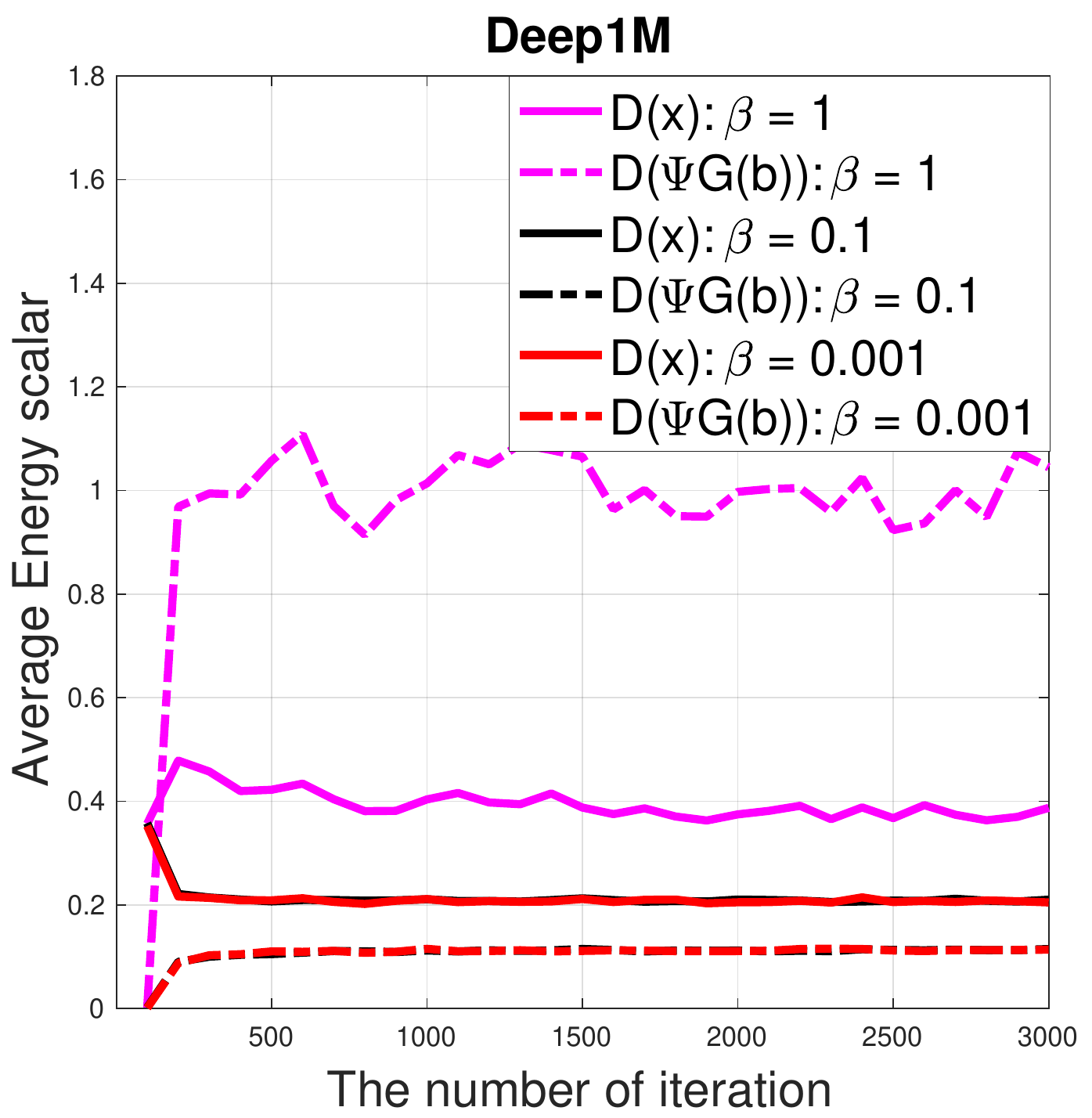}}\hspace*{-0.05\linewidth}
\subfigure[Energy vs. $\#$Iter.]{
\includegraphics[width=1.09\linewidth]{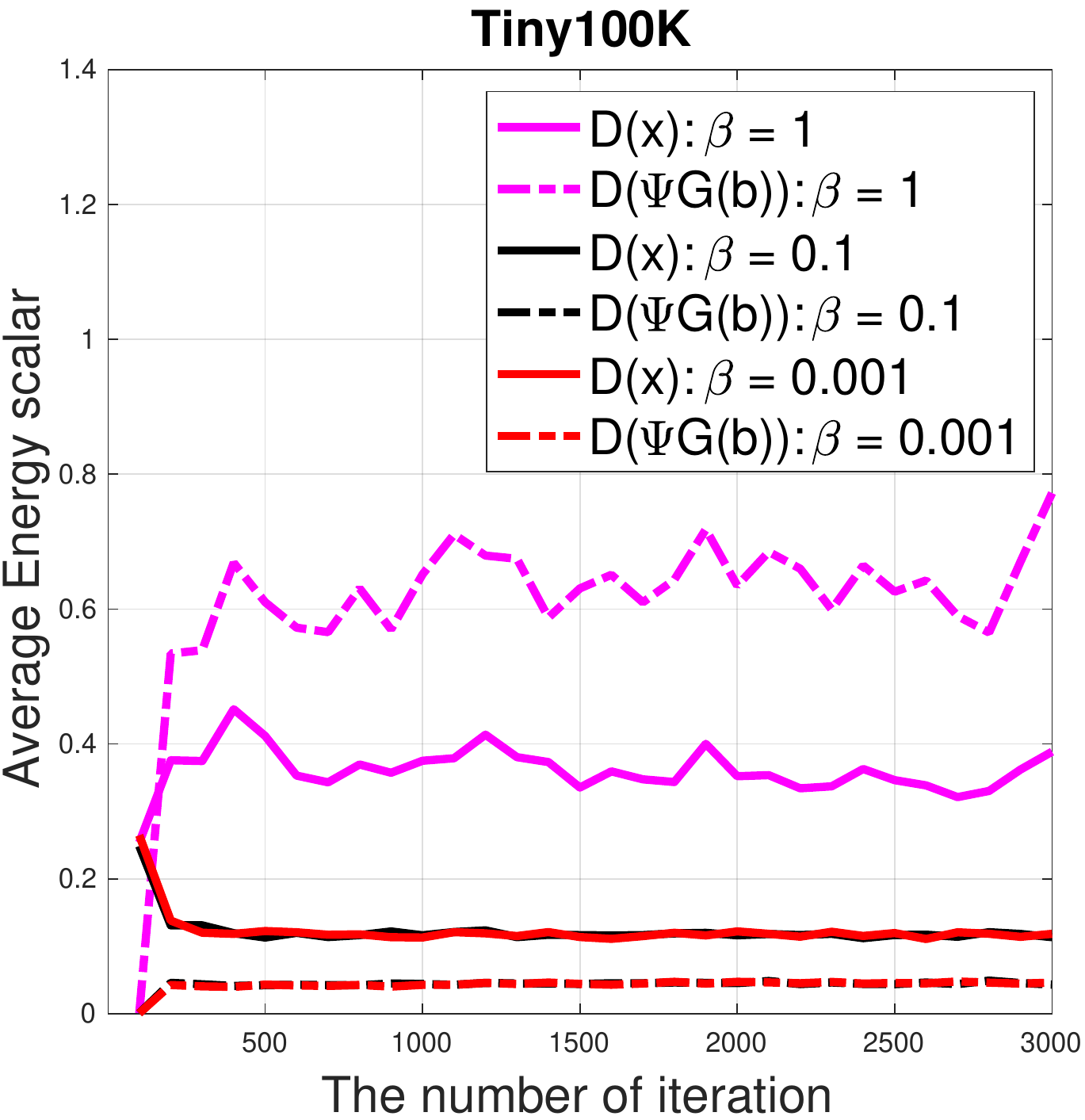}}\hspace*{-0.05\linewidth}
\subfigure[Reconstruction vs. $\#$Iter.]{
\includegraphics[width=1.1\linewidth]{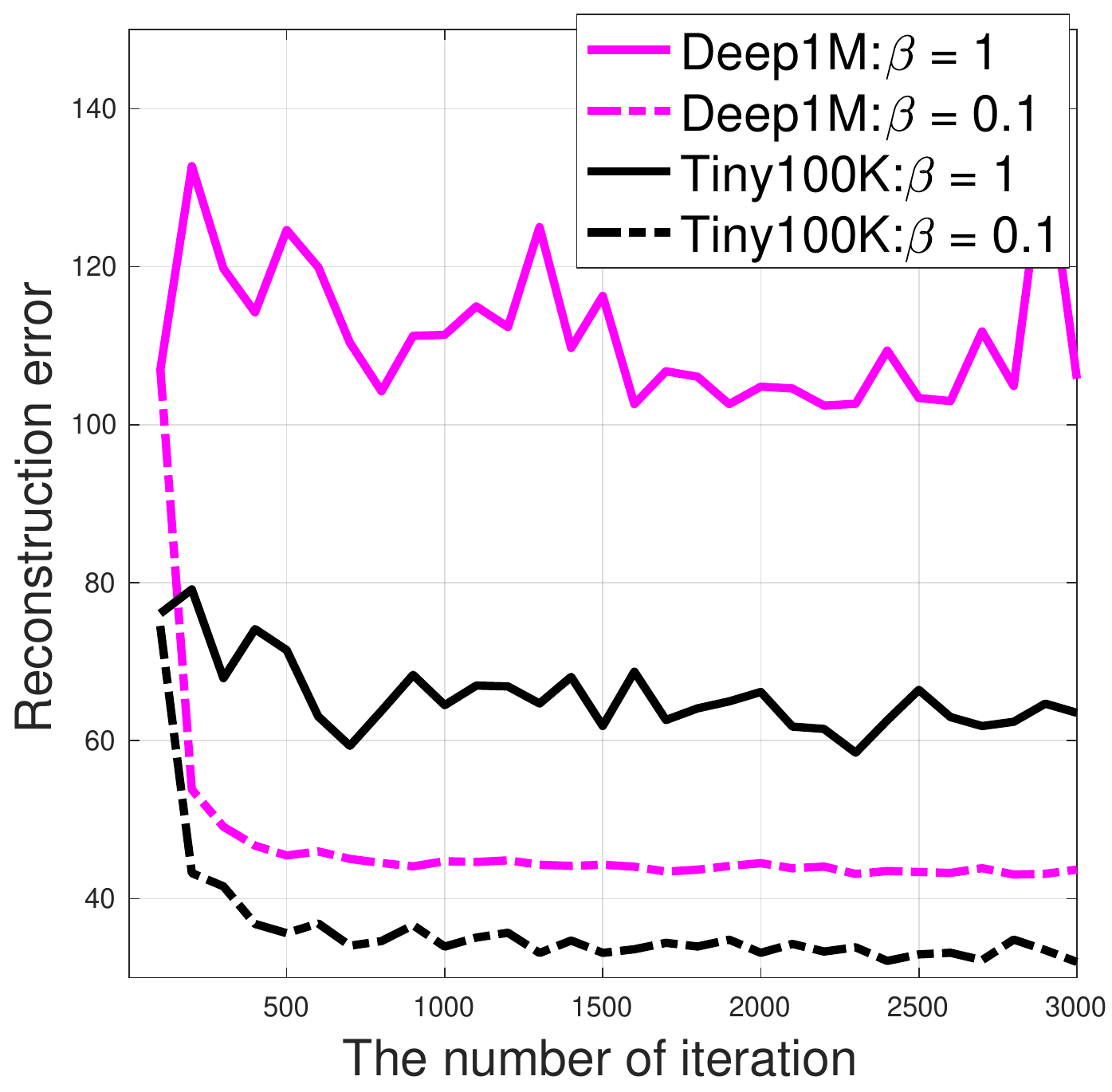}}
}
\end{minipage}
\end{center}
\vspace{-1.0em}
\caption{ Influence of the proposed energy model in Sec.\ref{sec32}. (Best viewed in color.)}
\label{fig7}
\end{figure*}

\subsection{Reconstruction Analysis} \label{sec44}

Based on the proposed SiGM, we can easily  reconstruct  an input vector through Eq.(\ref{eq4}).
The reconstruction error is computed using  the L$2$-norm loss of the reconstructed input and the original feature $\mbx$, as shown in Eq.(\ref{eq11}).
We first analyze the reconstruction error with different scales for the random matrix $\mbPsi$.
The matrix $\mbPsi$ can be seen as a random dictionary, which contains $m$ random dictionary vectors with $d$-dimension.
From the results shown in Fig.\ref{fig4} (a) and (c) on two different datasets for different retrieval tasks, a larger $m$ achieves less reconstruction error.
In Fig.\ref{fig_gen}, we also compare the reconstruction ability of different $m$ in SiGAH with those of BA and  SGH, when the length of the hash bit is $64$.
Then, compared to BA and SGH, SiGAH produces a better reconstruction quality, in which the edges of the number are more clear, and the shapes are more similar to the real image.
The reconstruction error of SiGAH is less than SGH,  which demonstrates our claim that more accurate reconstruction brings better ability of modeling data distribution.
Note that QoLSH with $784$ bits also achieves high reconstruction quality, but its background is very noisy.
In general, if $m$ is large enough, the dictionary is over-completed, so that the original feature can be sufficiently recovered by a sparse representation even if a random dictionary is used \cite{coates2011importance}.

Then, we compare the \emph{m}AP with different values for $m$, and the results are shown in Fig.\ref{fig4} (b) and (d).
We conclude that a larger $m$ can further improve not only the two retrieval tasks but also the quality of reconstructed images.
We also find that the performance plateaus after $m$ becomes twice the value of the input feature dimension.
As a result, $m = 2\times d$ is used as the default setting in all experiments.

\begin{table}[t]
\centering
\caption{Training time (s) comparison for different algorithms on three benchmark  datasets.}
\label{tab3}
\scalebox{1.0}[1.0]{
\begin{tabular}{c||c c | c c | c c}
\hline
              & \multicolumn{2}{c|}{GIST1M} & \multicolumn{2}{c|}{Deep1M} & \multicolumn{2}{c}{Tiny100K} \\ \hline\hline
Methods           & 32 bits         & 64 bits        & 32 bits          & 64 bits         & 32  bits          & 64  bits          \\ \hline
ITQ                     & 1.53       & 2.85       & 0.43        & 0.982        & 4.47          & 5.58          \\ 
BA                      & 534       & 1.12e3    & 163.7      & 225.2        & 514.8          & 1.35e3       \\ 
SGH                   & 320.7   & 350.4       & 201.6        & 230.2        & 301.8          & 340.4        \\ 
OCH                  & 35.3      & 37.5        & 6.74          & 9.08       & 19.9        & 23.01         \\ 
\textbf{SiGAH}  & 85.3    & 98.2      &  25.89       & 35.93       & 69.11         & 86.69         \\ \hline
\end{tabular}}
\end{table}

\subsection{Ablation Study}
The training time comparison is listed in Table \ref{tab3}. 
Our method (SiGAH) achieves competitive results with even less training time.
We discuss the influence of the proposed energy-based discriminative model of Eq.(\ref{eq19}) in Tab.\ref{tab_new} and Fig.\ref{fig7}.
First, we evaluate the importance of the adversarial learning in Table \ref{tab_new}.
We delete the discriminative part in the original SiGAH, and the model is transformed to an auto-encoder model to train the hash functions, making it similar to SGH \cite{Dai2017StochasticGH}.
Obviously, the adversarial learning can help to improve the retrieval performance over three benchmarks with an average gain in $m$AP of $20.9\%$.

\begin{table}[!t]
\centering
\caption{\emph{m}AP comparison using Hamming ranking on three benchmarks with and without adversarial learning. ``w/o'' means that we delete the discriminator in SiGAH, and ``w'' indicates the original SiGAH.}
\label{tab_new}
\scalebox{1.0}[1.0]{
\begin{tabular}{c||c|c|c|c|c|c}
\hline
    & \multicolumn{2}{c|}{GIST1M} & \multicolumn{2}{c|}{Deep1M} & \multicolumn{2}{c}{Tiny100K} \\ \hline
bit & 32           & 64           & 32           & 64           & 32            & 64            \\ \hline
w/o & 0.1976       & 0.2594       & 0.2024       & 0.3118       & 0.1576        & 0.2348        \\ 
w   & 0.2076       & 0.2594       & 0.2156       & 0.3307       & 0.1576        & 0.2425        \\ \hline
\end{tabular}}
\end{table}

\begin{figure*}[!t]
\begin{center}
\begin{minipage}[t]{0.23\linewidth}
\centerline{
\hspace*{-0.05\linewidth}
\subfigure[\small{\emph{m}AP vs.  $\alpha$}]{
\includegraphics[width=1.1\linewidth]{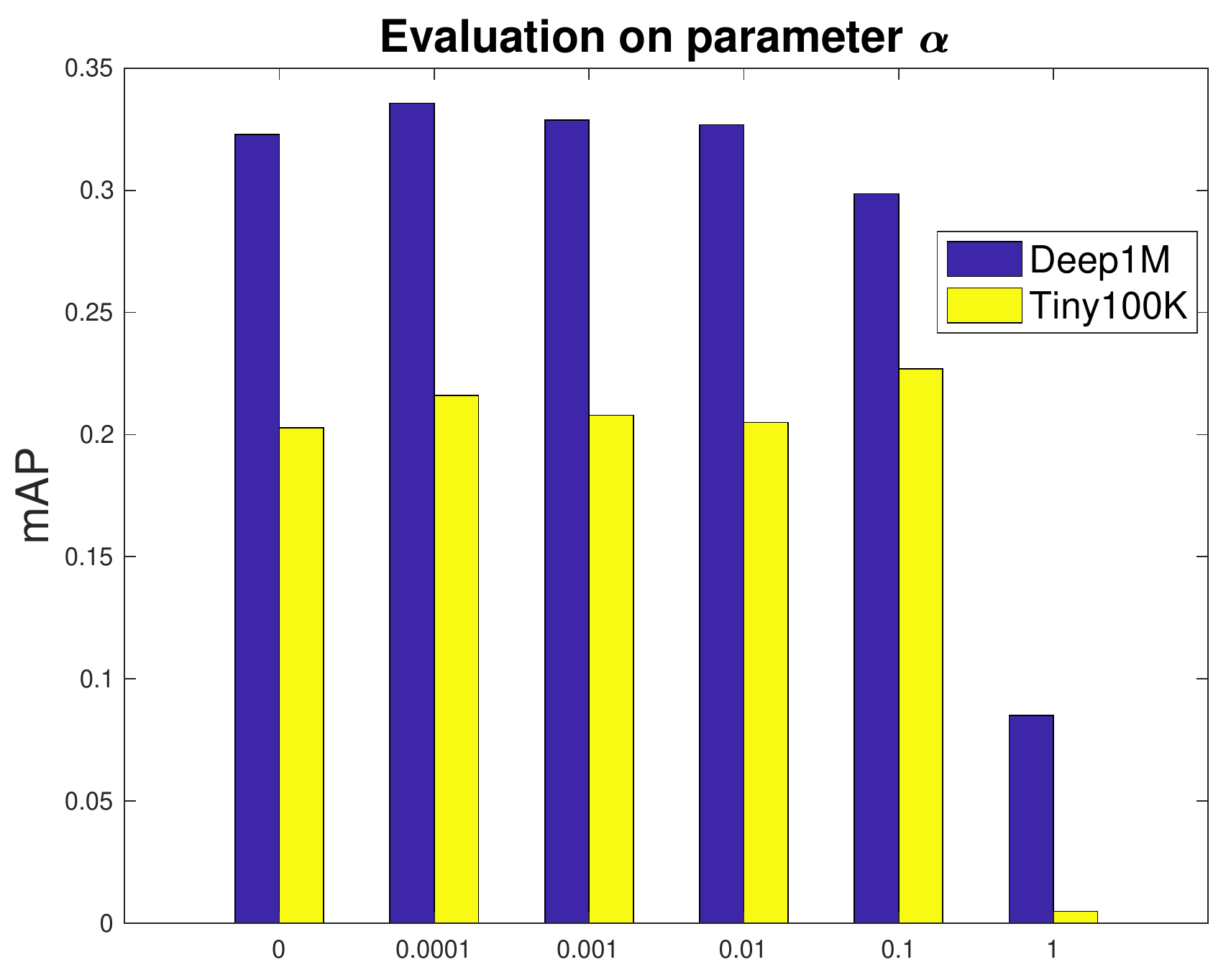}}\hspace*{-0.05\linewidth}
\subfigure[\small{\emph{m}AP vs.  $\gamma$}]{
\includegraphics[width=1.1\linewidth]{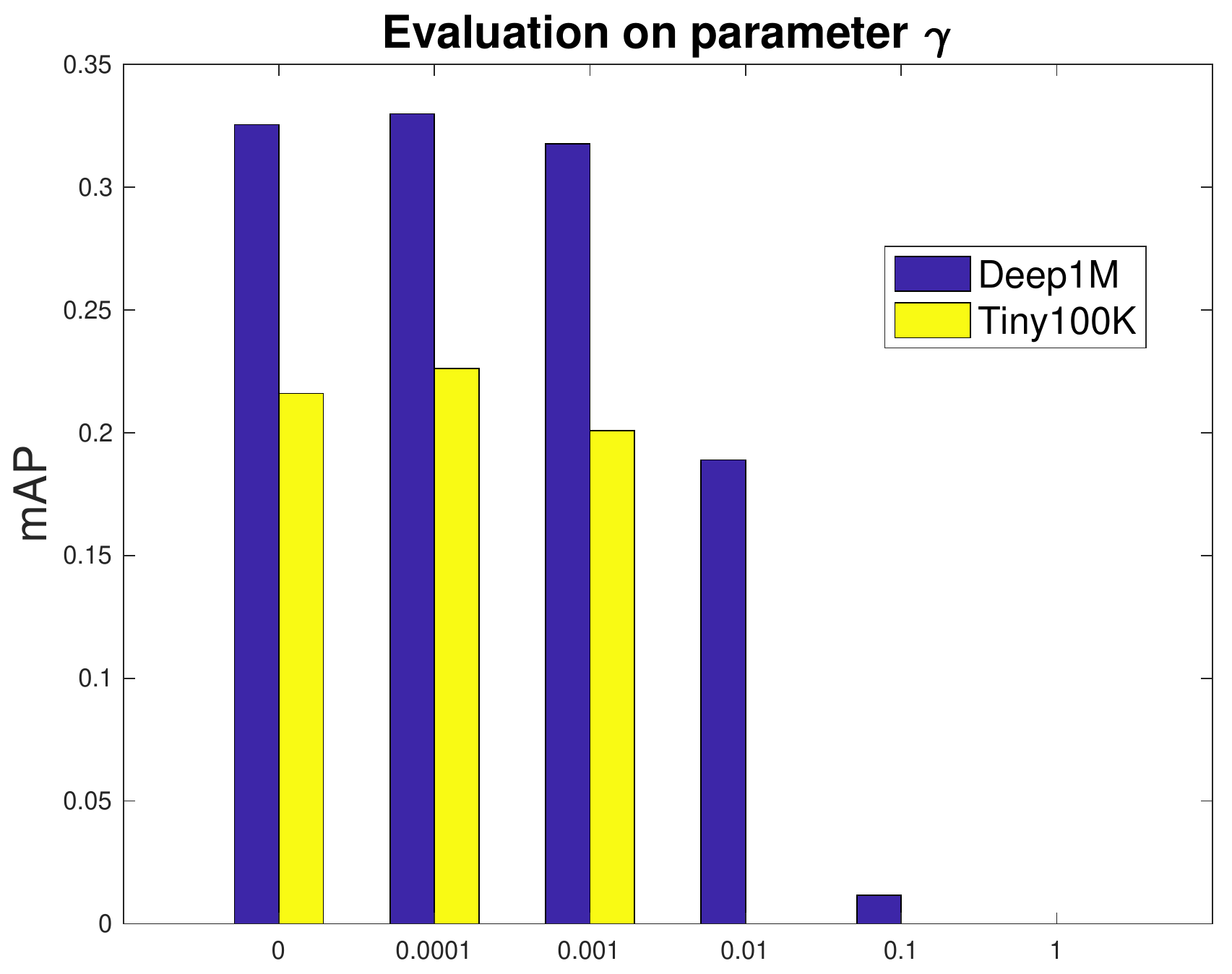}}\hspace*{-0.05\linewidth}
\subfigure[\small{\emph{m}AP vs.  $\lambda$}]{
\includegraphics[width=1.1\linewidth]{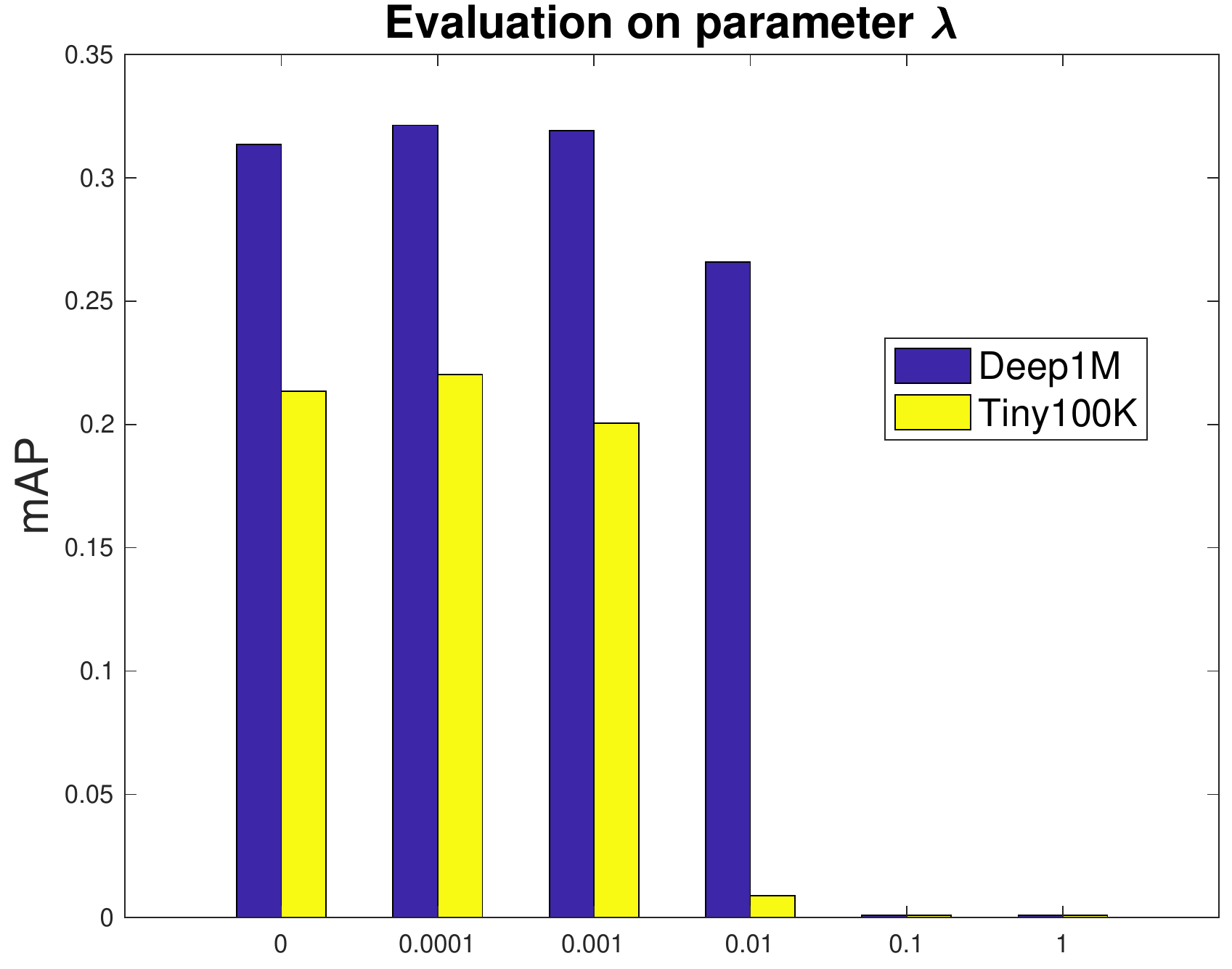}}\hspace*{-0.05\linewidth}
\subfigure[\small{Rec vs. Arch}]{
\includegraphics[width=1.05\linewidth]{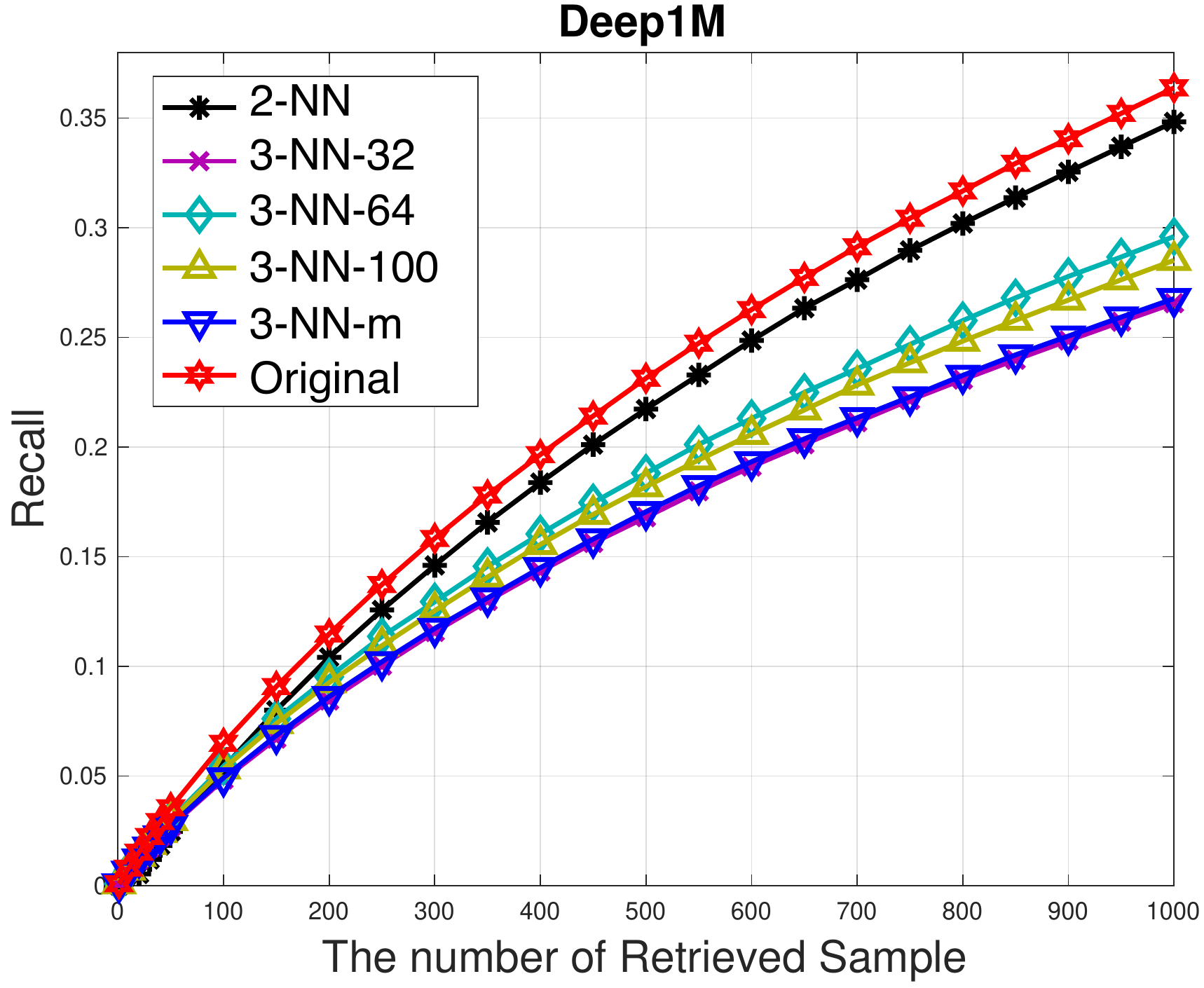}}
}
\end{minipage}
\end{center}
\vspace{-0.5em}
\caption{ The analysis of hyper-parameters and  different architecture of generative function $\FG$. (Best viewed in color.)}
\label{fig_last}
\end{figure*}
The relationship between \emph{m}AP and  the marginal parameter $\beta$ is  shown in Fig.\ref{fig7} (a).
This is conducted on Tiny100K, Deep1M, and MNIST. 
The \emph{m}AP score decreases quickly when $\beta$ is larger than $0.1$.
The relationship between the energy scalar and the number of iterations are shown in  Fig.\ref{fig7} (b) and (c). 
Although the discriminator gives lower energy to real data and higher energy to synthetic data, the difference between them is too large, which causes the generated feature to become outliers of the data manifold.
Furthermore, a larger $\beta$ will always result in a larger reconstruction error and prevent the model from converging, according to Fig.\ref{fig7} (d).
As such, a smaller $\beta$ in the discriminator can push the generative model to produce closer data energy scalars, which makes the distributions  close to each other and further improves the retrieval performance.
Comparing different $\beta$ on Tiny100K, Deep1M, and MNIST, the proposed SIGAH gains a competitive \emph{m}AP score when it is set to $0.1$, which is used as the  fixed parameter in the quantitative evaluations (Sec.\ref{sec43}).

\begin{table}[t]
\centering
\caption{The evaluation with different kinds of active function, which maintains $G(\cdot)$  in \emph{L}-Lipschitz families. }
\label{tab4}
\scalebox{1.0}[1.0]{
\begin{tabular}{c||c c c}
\hline
Method & Tiny100K & Deep1M & MNIST \\ \hline\hline
sigmoid    & \underline{0.2236} & \textbf{0.3325}  & - \\
tanh     & 0.2207 & 0.3305  & \underline{0.4872} \\ 
ReLU    & 0.2116 & 0.3079  & 0.4697  \\ 
LeakyReLU   & \textbf{0.2348} & \underline{0.3316}  & \textbf{0.5060}  \\ \hline
SWISH   & \underline{0.2327} & 0.3261  & \underline{0.4986}  \\ \hline
\end{tabular}}
\end{table}

We also analyze the performance of the proposed SiGM in Sec.\ref{sec31}.
To this end, we evaluate different active functions used in our generative model, which   mainly control the sparsity of the output $\FG$.
As mentioned in Sec.\ref{sec31}, the generative model should be ReLU-based Neural Network, so we compare three different active functions, \emph{i.e.}, sigmoid function (sigmoid), a hyperbolic tangent function (tanh), SWISH, and two ReLU-based functions (traditional ReLU, and LeakyReLU).
As shown in Table \ref{tab4}, the LeakyReLU activation achieves the best results on all three datasets, so we use LeakyReLU as our final activation function in all experiments.

We plot the \emph{m}AP score of the Hamming ranking with different values for the three hyper-parameters, \emph{i.e.}, $\alpha$, $\lambda$, and $\gamma$, in Fig.\ref{fig_last} (a)-(c).
We observe that \emph{m}AP decreases with a larger parameter value for all three parameters.
Although setting all the parameters to $0$ can also achieve competitive results, the best performance is achieved when they are varied from $1e-3$ to $1e-4$.
As a result, we empirically set  $\lambda$ and $\gamma$ to  $1e^{-4}$ and $1e^{-4}$, respectively.
We set $\alpha$ to $0.1$ on Tiny100K, and $1e^{-4}$ on the other datasets.

Finally, we replace the generator in Fig.\ref{fig_stru} with five different multi-layer neural networks with ReLU activations. 
For a fair comparison, we first apply a $4$-layer auto-encoder model (4-AE), where the dimensions for the layers are set as $[d-r-p-d]$ with four different values for $p$ (\emph{i.e.}, $32$, $64$, $100$, and $m$).  
We also consider a $3$-layer auto-encoder model (3-AE), where the dimensions for the layers are set as $[d-r-d]$.
The parameters in all five neural networks can be trained with the same learning rate and learning epochs. 
The quantitative results are shown in Fig.\ref{fig_last} (d), which fully explains the advantages of the proposed scheme.

\subsection{Comparison with State-of-the-Art Deep Hashing}
At last, in addition to the retrieval performances with traditional hashing, we also compare our SiGAH with recent representative unsupervised deep hashing \cite{Liong2015,8417979,Dai2017StochasticGH,su2018greedy} and GAN-based hashing \cite{ghasedi2018unsupervised,Maciej2018}.

We mainly report the retrieval performances on \textbf{CIFAR-10} and \textbf{ImageNet}  for a comprehensive view on encoding quality. 
\textbf{CIFAR-10} labeled subsets of the $80$ million tiny images dataset, which consists of $60,000$ $32 \times 32$ color images in $10$ classes.
We select 10,000 images (1,000 per class) as the query set, and the remaining 50,000 images are regarded as the training set.
\textbf{ImageNet} consists of 1,000 image classes for object classification and detection. 
Following \cite{su2018greedy}, we randomly select 100 categories to perform our retrieval task. All the original training images are used as the database, and all the validation images form the query set.
Moreover, 100 images per category are randomly selected from the database as the training points.
Following the similar setting in \cite{Maciej2018}, we report the \emph{m}AP of top $1,000$ returned images with respect to different number of hash bits in Table \ref{tabf}.
For fair comparisons, all the methods use identical training and test sets.

The first two lines in Table \ref{tabf} report the retrieval performance with the deep hashing (GIST with $512$ dimension), which aims to compare the SiGAH to classical deep hash (DeepBit) \cite{Liong2015}.
It is clear that SiGAH significant outperform the DH for different hash bits.
Note that SiGAH just use the linear hash function, but the DH use two-layer fully-connected layer to encode the input feature to binary codes.
This has demonstrated that SiGAH can learn better binary codes with simple linear hash function.


\begin{table}[]
\centering
\caption{Performance comparisons (w.r.t. mAP@1000) of SiGAH and the state-of-the-art deep hashing methods.}\label{tabf}
\scalebox{0.9}[0.9]{\begin{tabular}{c||ccc|ccc}
\hline
\multirow{2}{*}{Method} & \multicolumn{3}{c}{CIFAR-10} & \multicolumn{3}{c}{ImageNet} \\ \cline{2-7} 
                        & 16 bits  & 32 bits  & 64 bits & 16 bits  & 32 bits  & 64 bits \\ \hline
DeepBit                & 0.162    & 0.167    & 0.170   & -   & -    & -   \\ 
\textbf{SiGAH}                 & 0.214    & 0.240    & 0.254   & -    &-    & -   \\ \hline                        
ITQ-CNN                 & 0.385    & 0.414    & 0.442   & 0.217    & 0.317    & 0.391   \\ 
SpH-CNN                 & 0.302    & 0.356    & 0.392   & 0.185    & 0.271    & 0.350   \\ 
HashGAN                 & 0.447    & 0.463    & 0.481   & -        & -        & -       \\ 
BinGAN                 & 0.301    & 0.347    & 0.368   & -        & -        & -       \\ 
DeepBit                 & 0.194    & 0.249    & 0.277   & 0.204    & 0.281    & 0.286   \\ 
DBD-MQ                 & 0.215    & 0.265    & 0.319   & -   & -    & -   \\ 
SGH                     & {0.478}    & {0.512}    & 0.528   & \underline{0.447}    & 0.540    & 0.563   \\ 
GreedyHash              & 0.433    & 0.472    & \underline{0.521}   & 0.446    & \underline{0.577}    & \underline{0.698}   \\
TBH                 & \textbf{0.536}   & \underline{0.520}    & 0.512   & -   & -    & -   \\ 
\textbf{SiGAH(D)}                   & \underline{0.514}    & \textbf{0.525}    & \textbf{0.564}   & \textbf{0.533}    & \textbf{0.666}    & \textbf{0.758}   \\ \hline
\end{tabular}}
\end{table}

We further compare SiGAH to four unsupervised CNN-based hashing methods, such as DeepBit \cite{Lin2016},  DBD-MQ \cite{DBD}, SGH \cite{Dai2017StochasticGH}, and GreedyHash \cite{su2018greedy}, and two unsupervised GAN-based hashing methods, such as BinGAN \cite{Maciej2018} and HashGAN \cite{ghasedi2018unsupervised}. 
All these methods use CNN model with hash layer to encode binary codes for each image.
For fair comparison between our method and these deep models, we use the DeCAF feature \cite{donahue2014decaf}\footnote{Each image is represented by a $4,096$-dimensional deep feature from the fully-connected layer. The deep features are obtained via a 16-layer VGGNet \cite{Simonyan2014VeryDC}. The VGG-Net model was trained on the training set of ImageNet.} as inputs for SGH, GreedyHash and our SiGAH.

\begin{table}[]
\centering
\caption{Performance comparisons (w.r.t. mAP@1,000) under different CNN models, which is evaluated on the CIFAR-10.} \label{tlab}
\begin{tabular}{c||ccc}
\hline
          & \multicolumn{3}{c}{CIFAR-10}   \\ \hline
CNN Model & AlexNet  & VGG16     & VGG19    \\ 
SiGAH     & 0.461    & 0.564     & 0.573    \\ \hline
CNN Model & ResNet50 & ResNet152 & DenseNet121 \\ 
SiGAH     & 0.641    & 0.717     & 0.635    \\ \hline
\end{tabular}
\end{table}

The retrieval mAP@1,000 results of SiGAH are provided in Table \ref{tabf}. In fact, the deep features can significantly improve the retrieval performances, and we name SiGAH with deep feature as SiGAH(D).
The performance gap between SiGAH and existing unsupervised deep methods can be clearly observed. 
Particularly, SiGAH obtains remarkable average mAP gain about 11.54\% with different hash bits, when comparing to the second best method (as the underline score in Table \ref{tabf}).
Note that, recent deep hashing are always categorized into  supervised method, it is not reasonable to expect unsupervised hashing models to outperform all existing supervised ones.
However, under the same experimental setting on ImageNet we are followed \cite{su2018greedy}, SiGAH reaches the competitive performances of the supervised version of GreedyHash, as the Table 3 shown in \cite{su2018greedy}.
Moreover, SiGAH with hand-craft feature outperforms state-of-the-art unsupervised hashing methods, such as  DeepBit, when hash bit is $16$.
These two results show that the SiGAH with linear hash function and deep feature input also realizes better image retrieval under low hash bit, without any CNN's parameters' fine-tuning.

In Table \ref{tlab}, we also report the performances with different well-known CNN models, \emph{i.e.,} AlexNet \cite{Krizhevsky2012ImageNetCW}, VGG \cite{Simonyan2014VeryDC}, ResNet \cite{He:2016ib}, and DenseNet \cite{Huang:2019cv}.
As the results shown, SiGAH based on ResNet152 achieves the highest mAP score, that shows deeper CNN helps to improve the retrieval performance.
As a conclusion, SIGAH can produce high-quality binary codes, which is very efficient and effective when facing to the large-scale image retrieval.

\section{Conclusion} \label{sec6}
In this paper, we propose a novel unsupervised hashing method, termed Sparsity-Induced Generative Adversarial Hashing (SiGAH), which minimizes quantization error through reconstruction.
A generative adversarial framework is proposed to produce robust binary codes, which mainly contains a sparsity-induced generator and an MSE-loss based discriminator. 
The sparsity-induced generative model takes binary codes as input, and outputs robust synthetic features, with an energy function to discriminate the quality of the synthetic features.
Furthermore, we introduce a sparsity constraint into the generative model, which follows the compressive sensing theorem to enforce the reconstruction boundary. 
This generative adversarial learning can be trained using the SGD with Adam optimizer.  
The proposed SiGAH also has a higher training efficiency and lower storage cost, which is very suitable for large-scale visual search.
Experimental results on  a series of benchmarks have demonstrated our outstanding performance compared to the state-of-the-art unsupervised hashing methods.

\bibliographystyle{plain}
\bibliography{sample}

\end{document}